\documentclass[hidelinks,11pt,letterpaper]{article}
\usepackage[sectionbib,square]{natbib}
\usepackage[margin=1in]{geometry}
\usepackage{inconsolata}
 
\usepackage[T1]{fontenc}
\usepackage[utf8]{inputenc}
\usepackage{setspace}
\usepackage{microtype}
\usepackage{amsfonts,amsmath,amssymb,amsthm}
\usepackage{fontawesome}
\usepackage{booktabs}
\usepackage{subcaption}
\usepackage{wrapfig}
\usepackage{multicol}
\usepackage{pifont}
\usepackage[usenames,dvipsnames]{xcolor}
\usepackage[font={small,color=black,onehalfspacing}, labelfont={bf}]{caption}
\usepackage{graphicx}
\usepackage[export]{adjustbox}
\usepackage{csquotes}
\usepackage{xurl}
\usepackage{mwe}
\usepackage{enumitem}
\usepackage{titlesec}
\usepackage{palatino}
\usepackage{lipsum}
\usepackage{hyperref}       
\usepackage{nicefrac}       
\usepackage[english]{babel}
\usepackage{floatrow}

\newtheorem{problem}{Problem}
\newtheorem{proposition}{Proposition}
\newtheorem{lemma}{Lemma}

\newfloatcommand{capbtabbox}{table}[][\FBwidth]

\newenvironment{sciabstract}{%
\begin{quote}\small\onehalfspacing}
{\end{quote}}

\title{Expanding Knowledge Graphs with Humans in the Loop}

\author
{
  \begin{tabular}{c}
    Emaad Manzoor$^*$\\
    Cornell University\\
    \textit{emaadmanzoor@cornell.edu}\\[2em]
    Sriniketh Vijayaraghavan$^{*}$\\
    University of Wisconsin Madison\\
    \textit{svijayaragh3@wisc.edu}
  \end{tabular}
  \and
  \begin{tabular}{c}
    Jordan Tong$^{*}$\\
    University of Wisconsin Madison\\
    \textit{jordan.tong@wisc.edu}\\[2em]
    Rui Li\\
    Pinterest\\
    \textit{rli@pinterest.com}
  \end{tabular}
}

\date{}

\begin{document}

\maketitle
\thispagestyle{empty}

\onehalfspacing

\begin{center}
\textbf{Abstract}
\end{center}

\begin{sciabstract}
  Curated knowledge graphs encode domain expertise and improve the performance of recommendation, segmentation, ad targeting, and other machine learning systems in several domains.
  As new concepts emerge in a domain, knowledge graphs must be expanded
  to preserve machine learning performance.
  Manually expanding knowledge graphs, however, is infeasible at scale.
  In this work, we propose a method for knowledge graph expansion with humans-in-the-loop.
  Concretely, given a knowledge graph, our method predicts the ``parents''
  of new concepts to be added to this graph for further verification by human experts.
  We show that our method is both accurate and provably ``human-friendly''. Specifically, we prove that our method predicts parents that are ``near'' concepts' true parents in the knowledge graph, even when the predictions are incorrect. We then show, with a controlled experiment, that satisfying this property increases both the speed and the accuracy of the human-algorithm collaboration.
  We further evaluate our method on a knowledge graph from Pinterest and show that it outperforms competing methods on both accuracy and human-friendliness.
  Upon deployment in production at Pinterest, our method reduced the time needed for
  knowledge graph expansion by $\sim$400\% (compared to manual expansion), and contributed to a
  subsequent increase in ad revenue of $\sim$20\%.
\end{sciabstract}

\begin{quote}\small\textbf{Keywords:} Knowledge graphs, human-in-the-loop machine learning\end{quote}

{\let\thefootnote\relax\footnotetext{We are grateful to participants at the Cornell Artificial Intelligence seminar, the Cornell Marketing Brown Bag, the University of Southern California Marketing seminar, and the Artificial Intelligence in Management conference for their feedback. Part of this research was completed while Emaad Manzoor was at the Wisconsin School of Business. This research was supported in part by the University of Wisconsin Madison Office of the Vice Chancellor for Research and Graduate Education with funding from the Wisconsin Alumni Research Foundation. Computing support was provided in part by the Social Science Computing Cooperative at the University of Wisconsin Madison. Study participant recruitment support was provided by the BRITE Lab at the University of Wisconsin Madison.}}
{\let\thefootnote\relax\footnotetext{$^*$Authors contributed equally.}}

\clearpage

\doublespacing

\pagenumbering{arabic}

\section{Introduction}
\label{sec:introduction}



Knowledge graphs --- curated networks of concepts and entities such as consumers and merchants, product categories, travel destinations, and job skills --- are central to the ongoing wave of artificial intelligence that combines machine ``learning'' from data with computational ``reasoning'' \citep{deloitte2022kg}. By encoding human knowledge of a domain's concepts and their relationships in a machine-readable manner, knowledge graphs empower machine learning systems to generalize beyond the limited data they are trained on \citep{bahdanau2018systematic,wsdm2022yejin}.

Hence, firms across industries have invested in the construction of knowledge graphs to enhance the performance of various downstream machine learning systems \citep{forbes2021kg,hbr2022datagraphs}. For example, knowledge graphs are employed by Amazon to improve product recommendations \citep{dong2020autoknow,mao2020octet}, by Airbnb to link travelers to relevant Airbnb experiences \citep{airbnb2019kg}, by LinkedIn to connect job-seekers with suggested skills to acquire \citep{chen2018linkedin}, by Netflix to help content decision-makers \citep{netflix2020kg}, and by American Family Insurance to guide automated conversational agents \citep{bockhorst2019knowledge}.

Knowledge graphs are also increasingly being adopted for privacy-preserving ``interest-based advertising'' \citep{amazon2022taxonomy}. Key players such as the Google's Topics API \citep{lardinois_2022} and Pinterest's contextual targeting \citep{pinterest2022interesttargeting} rely on a hierarchical knowledge graph of ``interests'' that are used to categorize platform content. Interests then guide contextual ad targeting and real-time bidding, and the knowledge graph is used to broaden ads' reach by cross-targeting each interest's more-general ``parents'' in the graph \citep{pinterest2019expanded}.

As new concepts emerge in a domain, knowledge graphs must be expanded with new nodes to prevent potential revenue losses due to stale recommendations and no-longer-relevant ads, among other adverse outcomes. Due to their business-critical nature, knowledge graphs are typically expanded manually by teams of expert curators (taxonomists or ontologists). However, manual expansion does not scale with the rate of content creation on online platforms and marketplaces today. At Pinterest, for example, 8 curators recently expanded the knowledge graph by 6,000 new nodes over the course of a month, averaging 1.5 minutes per new node (by conservative work-hour estimates) \citep{gonccalves2019use}. This is in stark contrast with the millions of visual bookmarks created by Pinterest users every day, each potentially adding new concepts to the platform.

Hence, several algorithms have recently been proposed in the computer science literature to automate knowledge graph expansion \citep{yu2020steam,shen2020taxoexpan,liu2021temp,wang2021enquire,zhang2021taxonomy,zeng2021enhancing,shen2022automated}. These algorithms predict, for each concept to be added to a knowledge graph, which other nodes in the knowledge graph the new concept should be linked to. However, all of these algorithms focus on maximizing prediction accuracy assuming fully-automated knowledge graph expansion,
and ignore their (more likely) role as decision support systems to be used with humans-in-the-loop.

Recent research in several fields --- information systems, operations, judgement and decision-making, computer science, and economics, among others --- stresses that building performant decision support algorithms requires accounting for how they interact with humans over and above their accuracy \citep{kleinberg2018human,ludwig2021fragile,malik2020does,kim2022err,donahue2022human}.
For example, prior work finds that decision support algorithms benefit from incorporating human judgement and feedback \citep{ijcai2021p237,ibrahim2021eliciting,de2022doubting}, from capturing how humans deviate from the algorithms' predictions \citep{bastani2021improving,grand2022best,wolczynski2022learning}, and by optimizing the relative allocation of human and algorithmic effort \citep{raghu2019algorithmic,fugener2022cognitive}.

Motivated by this research, we propose a knowledge graph expansion method that is both accurate and that provably limits the effort required by humans-in-the-loop to fix its predictions when they are incorrect --- a property we call \textit{human-friendliness}. We derive this property by extending the triplet loss \citep{schroff2015facenet} minimized by our proposed method to be ``graph-aware''. We then theoretically upper-bound the expected network distance between the predicted and ``true'' nodes to be linked to new concepts being added to the knowledge graph. This expected network distance is essentially a graph-theoretic proxy for human effort; for incorrect predictions, humans need only probe a bounded network neighborhood around the predicted nodes to find the ``true'' nodes to be linked to that concept, without having to search the entire knowledge graph.

While we expect human-friendliness to improve the performance of the human-algorithm collaboration, similar human-centric properties have been shown to reduce performance by increasing humans' ``blind'' reliance on algorithms' potentially-incorrect predictions \citep{smith2020no,ahn2021will}. Hence, we design a controlled experiment to test how human-friendliness impacts the performance of knowledge graph expansion with humans-in-the-loop.

We find that human-friendliness significantly improves human-algorithm collaboration performance. Experimental subjects who were provided human-friendly decision support scored 97\% more on our knowledge graph expansion task than subjects who were provided not-human-friendly decision support, which we attribute to an increase in their decision-making speed and accuracy (by 21\% and 15\%, respectively). We also find that human-friendliness is particularly impactful when the decision support is incorrect. In such cases, subjects provided human-friendly decision support scored 800\% more, were 34\% faster, and were 42\% more accurate. While human-friendliness does directionally increase subjects' mistakes due to ``blind'' reliance on the decision support, the negative impact on performance is far outweighed by the speed and accuracy improvements.

We then conduct a case study in partnership with Pinterest --- a visual bookmarking platform with over 400 million monthly active users --- to evaluate our proposed method on a knowledge graph from the field. The knowledge graph we benchmark on is a human-curated network of ``interests'' that is used to enhance search results, recommendations, and targeted ads across the Pinterest platform. We show that our proposed method outperforms 8 competing methods on both accuracy and human-friendliness metrics. We further show that competing methods (unlike our proposed method) may exhibit a tradeoff between these two metrics, and understanding why this tradeoff occurs is difficult due to competing methods' lack of theoretical guarantees.

Our proposed method was successfully deployed to expand the Pinterest knowledge graph from $\sim$11,000 to $\sim$28,000 interests. Using our proposed method, knowledge graph curators spent $\sim$400\% less time per new interest added on average (relative to prior manual expansion efforts). Subsequently, an advertising signal called ``pin2interest'' based on the expanded knowledge graph \citep{pinterest2019pin2interest} improved the revenue of Pinterest Shopping ads \citep{pinterest2022shopping} by $\sim$20\% (relative to the revenue before the knowledge graph expansion).

\textbf{Contributions.} Our contributions are four-fold. First, we encode a desirable property of human-algorithm collaborations --- the ease of fixing algorithms' incorrect predictions --- into the loss function of a machine learning method without sacrificing its accuracy. Second, we theoretically guarantee that this desirable property is satisfied in expectation. Third, we experimentally validate that satisfying this property improves the performance of the human-algorithm collaboration, and further explain why. And fourth, we benchmark against a suite of competing methods on a knowledge graph from the field and demonstrate superior performance on various metrics.

More generally, our work operationalizes a call by human-computer interaction practitioners for human-in-the-loop algorithms to ``provide paths forward from failure'' \citep{google2022pair}, in the context of knowledge graph expansion. Akin to fairness and interpretability, our work introduces a new human-centric dimension of methodological improvement: improving humans' ability to fix incorrect predictions. Particularly on problems where accuracy gains are at an impasse, our work suggests an alternative avenue for methodological progress.

\textbf{Outline.} We begin by defining the knowledge graph expansion problem in Section \ref{sec:problem}. We then introduce our proposed method in Section \ref{sec:algorithm} and show that it is provably ``human-friendly''. In Section \ref{sec:experiment}, we experimentally test whether and how human-friendliness impacts the performance of knowledge graph expansion with humans-in-the-loop. In Section \ref{sec:casestudy}, we benchmark our proposed method against a suite of state-of-the-art methods in a case study on expanding the Pinterest knowledge graph. We discuss related work in Section \ref{sec:related}, and conclude in Section \ref{sec:conclusion} with a summary of our contributions, findings, and their future research implications.

\section{Problem Definition}
\label{sec:problem}

Denote by $\mathcal{G} = (V, E)$ the knowledge graph to be expanded, represented as a directed acyclic graph with nodes $V$ and edges $E$. Each edge $(u, v) \in E$ represents a relationship between a \textit{child} node $u \in V$ and its \textit{parent} node $v \in V$. We assume that each node $u \in V$ is associated with a feature vector $\mathbf{e}_u \in \mathbb{R}^d$, derived using an external procedure that is independent of the knowledge graph. For example, the feature vectors could be word embeddings \citep{mikolov2013distributed,pennington2014glove} in a textual knowledge graph, image descriptors \citep{lowe1999object,bay2006surf} in a visual knowledge graph, or a combination of the two in a multimodal knowledge graph \citep{liu2019mmkg}.

\clearpage

Let $q \notin V$ be a \textit{query} node that is not (yet) part of the knowledge graph, and $\mathbf{e}_q \in \mathbb{R}^d$ be its feature vector (derived using the same external procedure used for nodes $u \in V$). Our goal is to predict the parent(s) of $q$ for further verification by human experts, who will subsequently attach $q$ to its ``true'' parent(s) in the knowledge graph\footnote{While it is possible to address cases when the knowledge graph has no suitable parents for a query node by adding a dummy ``no suitable parent'' node to the knowledge graph; we do not consider such cases in our current work.}. We formalize this as the following problem:
\begin{problem}[Knowledge Graph Expansion]
  Given a knowledge graph $\mathcal{G} = (V, E)$, feature vectors $\mathbf{e}_u \in \mathbb{R}^d$ for each node $u \in V$ and a query node $q \notin V$ with feature vector $\mathbf{e}_q \in \mathbb{R}^d$, rank the knowledge graph nodes such that the true parent(s) of $q$ are ranked higher than its non-parents.
  \label{problem1}
\end{problem}
Given a ranking of potential parents produced by a method that addresses Problem \ref{problem1}, humans-in-the-loop can use all or part of this ranking to attach the query node to its true parent(s). Though we focus on attaching each query node to its parent(s), it is straightforward to modify our problem to rank the potential child(ren) of each query node instead. It is also straightforward to extend our problem to \textit{insert} a query node between two nodes in the knowledge graph by predicting both its parent(s) and child(ren).

\section{Proposed Method}
\label{sec:algorithm}


Our proposed method is motivated by two goals. First, we want to produce an accurate ranking of potential parent(s) for each query node (the \textit{ranking quality goal}). Second, we want to help humans-in-the-loop find the true parent(s) of each query node even when the produced ranking is inaccurate (the \textit{human friendliness goal}). We hypothesize that achieving human friendliness, in addition to ranking quality, will improve the performance of the human-algorithm collaboration (we validate this in Section \ref{sec:experiment}). In this section, we propose a method to satisfy both goals.

\subsection{Learning to rank node pairs}
\label{sec:learningtorank}

We propose learning a \textit{score function} $s(u,v | \Theta)$ parameterized by $\Theta$ to score pairs of nodes $(u, v)$ having feature vectors $\mathbf{e}_u$ and $\mathbf{e}_v$ respectively, such that $s(u,v | \Theta)$ is high when $(u, v)$ is a child-parent pair and $s(u,v | \Theta)$ is low otherwise. Given a score function $s(u, v|\hat{\Theta})$ learned from a training knowledge graph $\mathcal{G} = (V, E)$ and a query node $q \notin V$, we rank the knowledge graph nodes $v \in V$ in decreasing order of the scores $s(q, v|\hat{\Theta})$ to address Problem \ref{problem1}.

\textbf{Learning objective.} To learn $s(u,v | \Theta)$ from a training knowledge graph $\mathcal{G} = (V, E)$, we propose finding $\Theta$ such that the score of each training child-parent pair $(u,v) \in E$ is greater than the score of each non-child-parent node pair $(u, v') \notin E$ by a minimum margin. Concretely, we seek to find $\Theta$ that satisfies the following minimum-margin constraint for every child-parent pair $(u,v) \in E$:
\begin{align}
  s(u, v| \Theta) \geq s(u, v'| \Theta) + \gamma(u,v,v') \qquad \forall v' \in V - H(u),
  \label{eq:constraint}
\end{align}
where $H(u) \subset V$ is the set of parents of $u$, $v' \in V - H(u)$ is \textit{not} a parent of $u$, and $\gamma(u,v,v')$ is the desired minimum margin for the triplet $(u,v,v')$.

However, it may not be possible to find a $\Theta$ satisfying the minimum-margin constraint in Equation \ref{eq:constraint} for \textit{all} triplets of nodes $(u, v, v')$. Hence, we derive a \textit{minimum-margin loss function} motivated by the constraint in Equation \ref{eq:constraint}. Minimizing this loss function corresponds to finding a $\Theta$ that satisfies the minimum-margin constraint in Equation \ref{eq:constraint} to the maximum extent possible.

Let $\mathcal{E}(u, v, v'| \Theta)$ be the extent to which a node $v' \in V$ that is not a parent of $u$ violates the minimum-margin constraint in Equation \ref{eq:constraint} for a child-parent pair $(u, v) \in E$:
\begin{align}
  \mathcal{E}(u, v, v'| \Theta) = \textrm{max}[0, s(u,v'| \Theta) - s(u, v| \Theta) + \gamma(u,v,v')].
  \label{eq:violation}
\end{align}
If the minimum-margin constraint is satisfied for the triplet $(u, v, v')$, then $\mathcal{E}(u, v, v'| \Theta) = 0$. Otherwise, $\mathcal{E}(u, v, v'| \Theta) > 0$. This is similar to the hinge loss used in support vector machines.

Our proposed minimum-margin loss function $\mathcal{L}(\mathcal{G}| \Theta)$ is the sum of the violations of the minimum-margin constraints over all triplets of nodes $(u, v, v')$:
\begin{align}
  \mathcal{L}(\mathcal{G}| \Theta) = \sum_{(u, v) \in E} \sum_{v' \in V - H(u)} \mathcal{E}(u, v, v'| \Theta)
  \label{eq:loss}
\end{align}
where, as in Equation \ref{eq:constraint}, $H(u) \subset V$ is the set of parents of $u$ and $v' \in V$ is not a parent of $u$. We then set $\hat{\Theta}$ to the minimizer of the minimum-margin loss function in Equation \ref{eq:loss}.

\textbf{Designing the score function $s(u,v|\Theta)$.} Note that we do not make any assumptions on the structure of the score function $s(u,v|\Theta)$. In practice, the score function should be designed based on the knowledge graph being expanded and encode domain-specific assumptions. If $s(u,v|\Theta)$ is designed to be a differentiable function of $\Theta$, then the minimizer $\hat{\Theta}$ of Equation \ref{eq:loss} can be found using computationally efficient gradient-based optimization techniques. 
In Section \ref{sec:relatedness}, we design a differentiable score function for our case study on the Pinterest knowledge graph.

\textbf{Connection to the triplet loss.} The loss function in Equation \ref{eq:loss} is a generalization of the triplet loss \citep{weinberger2009distance,schroff2015facenet}, which is widely used in machine learning for information retrieval and ranking. As such, one can design a score function inspired by this literature to produce high-quality rankings. However, such a score function would ignore the human friendliness goal. In the next subsection, we propose a modification of the loss function in Equation \ref{eq:loss} that theoretically guarantees human friendliness.

\subsection{Guaranteeing Human Friendliness with Knowledge Graph-Aware Margins}
\label{sec:theory}

The loss function in Equation \ref{eq:loss} includes a hyperparameter $\gamma(u,v,v')$, which is the desired minimum margin between the scores of child-parent and non-child-parent node pairs in the training knowledge graph. In prior work employing the triplet loss, $\gamma(u,v,v')$ is typically set to a constant $\gamma$ for all triplets $(u,v,v')$ \citep{schroff2015facenet}. Some recent approaches use a different $\gamma(u,v,v')$ for each triplet, set using heuristics \citep{wang2018deep} or learned from the data \citep{feng2019triplet}.

\textbf{Proposed margin.} In contrast with prior work, we propose using a margin that varies for each triplet $(u,v,v')$, requires no additional learning, and does not rely on heuristics. Concretely, for each child-parent pair $(u,v) \in E$ and node $v' \in V - H(u)$ that is not a parent of $u$, we set the minimum margin to the undirected shortest path distance\footnote{The undirected shortest path distance between two nodes in a graph is the minimum number of intermediate nodes or ``hops'' on a path connecting the two nodes. It is an integer ranging from zero to the \textit{diameter} of the graph. It is undefined for ``disconnected'' node pairs that do not have a path connecting them in the graph.} $d_{\mathcal{G}}(v,v')$ between $v$ and $v'$ in the graph $\mathcal{G}$:
\begin{align}
  \gamma(u, v, v') \equiv d_{\mathcal{G}}(v,v')
\end{align}
Our proposed margin is \textit{knowledge graph-aware}. If a parent $v$ and non-parent $v'$ of $u$ are ``far away'' in the knowledge graph, our proposed margin encourages the corresponding scores $s(u,v|\Theta)$ and $s(u,v'|\Theta)$ to be dissimilar. If a parent $v$ and non-parent $v'$ of $u$ are ``nearby'' in the knowledge graph, our proposed margin allows the corresponding scores $s(u,v|\Theta)$ and $s(u,v'|\Theta)$ to be similar.

\clearpage

\textbf{Guaranteeing human friendliness.} We now show that for any child-parent node pair $(u,v)$ (where $v \in V$ is in the knowledge graph but $u$ may not be in the knowledge graph), our proposed margin enables bounding the expected undirected shortest path distance between the true parent $v \in V$ and the \textit{top-ranked predicted parent} $\hat{v}(u) \in V$ of $u$. Hence, we can guarantee that the true and top-ranked predicted parents of nodes to be added to the knowledge graph will be ``nearby'' in the knowledge graph in expectation, even when the predictions are inaccurate.

Our first theoretical result bounds the expected undirected shortest path distance between the true and top-ranked predicted parents of \textit{in-sample} nodes $u \in V$:
\begin{proposition}
  Let $(u_1, v_1), \dots, (u_n, v_n) \in E$ be the child-parent node pairs in the training knowledge graph $\mathcal{G} = (V,E)$. Then, the expected undirected shortest path distance between the true parent $v_i$ and top-ranked predicted parent $\hat{v}(u_i)$ of a training node $u_i \in V$ is bounded above by the expected training loss:
  \begin{align*}
     \frac{1}{n}\sum_{i=1}^n d_{\mathcal{G}}(v_i, \hat{v}(u_i)) \leq \frac{\mathcal{L}(\mathcal{G}|\hat{\Theta})}{n}
  \end{align*}
\end{proposition}
\begin{proof}
The top-ranked predicted parent of any node $u$ is the node $v \in V$ in the knowledge graph with the highest score $s(u,v|\hat{\Theta})$. The following inequality thus follows:
  \begin{eqnarray}
    s(u, \hat{v}(u)|\hat{\Theta}) - s(u, \tilde{v}|\hat{\Theta}) \geq 0 \quad \forall \tilde{v} \in V.
    \label{eq:theorem-inequality-1}
  \end{eqnarray}
  Since $\gamma(u, v, \hat{v}(u)) = d_{\mathcal{G}}(v, \hat{v}(u))$, we can use equations (\ref{eq:violation}) and (\ref{eq:theorem-inequality-1}) to lower-bound the minimum-margin constraint violation of the top-ranked predicted parent as follows:
  \begin{eqnarray}
    \mathcal{E}(u, v, \hat{v}(u)|\hat{\Theta}) &=& \textrm{max}[0, s(u,\hat{v}(u)|\hat{\Theta}) - s(u, v|\hat{\Theta}) + \gamma(u, v, \hat{v}(u))]\nonumber\\
                                  &\geq& s(u, \hat{v}(u)|\hat{\Theta}) - s(u, v|\hat{\Theta}) + \gamma(u, v, \hat{v}(u))\nonumber\\
                                  &\geq& \gamma(u, v, \hat{v}(u)) = d_\mathcal{G}(v, \hat{v}(u))
  \end{eqnarray}
  \clearpage
  Using the fact that $\mathcal{E}(u, v, v'|\Theta) \geq 0$, and summing over the training child-parent node pairs $(u_1, v_1), \dots, (u_n, v_n)$ and the corresponding nodes $v'$ that are not parents of $u_i$ concludes our proof:
  \begin{eqnarray}
    \mathcal{L}(\mathcal{G}|\hat{\Theta}) &=& \sum_{i=1}^n \sum_{v' \in V - H(u_i)} \mathcal{E}(u_i, v_i, v'|\hat{\Theta})\\
                            &\geq& \sum_{i=1}^n \mathcal{E}(u_i, v_i, \hat{v}(u_i)|\hat{\Theta})\nonumber\\
                            &\geq& \sum_{i=1}^n d_\mathcal{G}(v_i, \hat{v}(u_i))\nonumber\\
    \implies \frac{\mathcal{L}(\mathcal{G}|\hat{\Theta})}{n} &\geq& \frac{1}{n}\sum_{i=1}^n d_\mathcal{G}(v_i, \hat{v}(u_i))\nonumber
  \end{eqnarray}
\end{proof}
\textbf{Implications of Proposition 1.} The training loss is minimized at $\hat{\Theta}$, which parameterizes the score function $s(u,v|\Theta)$ used to rank the potential parents of each node $u$ to be added to the knowledge graph. Hence, if the training loss is successfully\footnote{The success of minimization depends on properties of the score function and on the optimization technique used to minimize the minimum-margin loss in Equation (\ref{eq:loss}).} minimized to zero or a small number, Proposition 1 guarantees that the true and top-ranked predicted parent(s) of in-sample nodes $u \in V$ will be ``nearby'' in terms of their undirected shortest path distance in $\mathcal{G}$.

In practice, our proposed method will be used to rank the potential parents of \textit{out-of-sample} nodes $u \notin V$. This motivates our second theoretical result, which relies on the following lemma:
\begin{lemma}\citep{hoeffding1963probability} Let $X_1, \dots, X_n \sim \mathcal{D}$ be independent random variables drawn from some distribution $\mathcal{D}$ with expected value $\mathbb{E}_{X \sim \mathcal{D}}[X]=\mu$ and bounded range $-\infty < a \leq X_i \leq b < \infty$ for each $i=1,\dots,n$. Additionally, let $\bar{X}_n = n^{-1}\sum_{i=1}^n X_i$ be their empirical average. Then, for any $\epsilon > 0$:
  \begin{align}
    \mathbb{P}[|\bar{X}_n - \mu| 
    \geq \epsilon] \leq 2e^{-\frac{2n\epsilon^2}{(b-a)^2}}
  \end{align}
  \vspace{-11mm}
\end{lemma}
Lemma 1 (also called Hoeffding's Inequality) bounds the deviation between the expected value and the empirical average of a collection of independent and bounded random variables. Since the undirected shortest path distance between any pair of nodes in a graph is bounded between zero and the diameter of the graph, we use Lemma 1 to bound the expected undirected shortest path distance between the true and top-ranked predicted parents of out-of-sample nodes.

\begin{proposition}
  Let $u_1, \dots, u_n$ be the nodes of a training knowledge graph $\mathcal{G} = (V,E)$ with diameter $D$ independently drawn from some distribution $\mathcal{V}$, and let $v_1, \dots, v_n$ be their corresponding parents. Let $\mathcal{L}(\mathcal{G}|\hat{\Theta})$ be the resulting training loss after learning $\hat{\Theta}$ from $\mathcal{G}$. Further, let $u$ be any node drawn from the same distribution $\mathcal{V}$ (possibly out-of-sample). Then, with probability at least $1 - \delta$:
  \begin{align*}
    \mathbb{E}[d_{\mathcal{G}}(v,\hat{v}(u))] \leq \frac{\mathcal{L}(\mathcal{G}|\hat{\Theta})}{n} + D\sqrt{\frac{1}{2n}\textrm{ln}\frac{2}{\delta}}
  \end{align*}
\end{proposition}
\begin{proof}
  Since the training nodes $u_1, \dots, u_n$ are independent and identically distributed, for a given $\hat{\Theta}$, the undirected shortest path distances $d_{\mathcal{G}}(v_i, \hat{v}(u_i)), \dots, d_{\mathcal{G}}(v_n, \hat{v}(u_n))$ between the respective parents $v_1, \dots, v_n$ and the respective top-ranked predicted parents $\hat{v}(u_1), \dots, \hat{v}(u_n)$ are also independent and identically distributed. Further, each of these undirected shortest path distances is bounded between zero and the diameter $D$ of the training knowledge graph $\mathcal{G}$.
  
  Hence, we can apply Lemma 1 to  $d_{\mathcal{G}}(v_i, \hat{v}(u_i)), \dots, d_{\mathcal{G}}(v_n, \hat{v}(u_n))$. Let $\mathbb{E}[d_{\mathcal{G}}(v,\hat{v}(u))]$ be the expected undirected shortest path distance between the true parent $v$ and top-ranked predicted parent $\hat{v}(u)$ of a node $u$ drawn from the distribution $\mathcal{V}$. Then, for any $\epsilon > 0$:
  \begin{align}
    \mathbb{P}\left[\left|\frac{1}{{n}}\sum_{i=1}^n d_{\mathcal{G}}(v_i, \hat{v}(u_i)) - \mathbb{E}[d_{\mathcal{G}}(v,\hat{v}(u))]\right| 
    \geq \epsilon\right]
    &\leq 2e^{-\frac{2n\epsilon^2}{D^2}}
  \end{align}
  We can rewrite the inequality above as:
  \begin{align}
    \mathbb{P}\left[\left|\frac{1}{{n}}\sum_{i=1}^n d_{\mathcal{G}}(v_i, \hat{v}(u_i)) - \mathbb{E}[d_{\mathcal{G}}(v,\hat{v}(u))]\right| 
    \leq \epsilon\right]
    &\geq 1 - 2e^{-\frac{2n\epsilon^2}{D^2}}
  \end{align}
  Set $\delta = 2e^{-\frac{2n\epsilon^2}{D^2}}$, so $\epsilon = D\sqrt{\frac{1}{2n}\textrm{ln}\frac{2}{\delta}}$. Then, with probability at least $1 - \delta$:
  \begin{align}
    \mathbb{E}[d_{\mathcal{G}}(v,\hat{v}(u))] \leq \frac{1}{{n}}\sum_{i=1}^n d_{\mathcal{G}}(v_i, \hat{v}(u_i)) + D\sqrt{\frac{1}{2n}\textrm{ln}\frac{2}{\delta}}
  \end{align}
\end{proof}
\noindent Applying Proposition 1 to bound $n^{-1}\sum_{i=1}^n d_{\mathcal{G}}(v_i, \hat{v}(u_i))$ from above concludes our proof$\qedhere$.

Proposition 2 theoretically illuminates several factors that influence how ``nearby'' the true and top-ranked predicted parents of in-sample or out-of-sample nodes will be in expectation. Concretely, the upper bound on the expected undirected shortest path distance between the true and top-ranked predicted parents: (i) decreases with a lower training loss $n^{-1}\mathcal{L}(\mathcal{G}|\hat{\Theta})$, (ii) decreases with a larger training dataset size $n$, (iii) increases with the diameter $D$ of the knowledge graph, and (iv) increases with a higher desired probability $1 - \delta$. Hence, to obtain a meaningful upper bound, the number of training nodes $n$ should be large compared to the diameter of the graph $D$, and the training loss needs to be minimized to zero or a small value.

\textbf{Implications of Proposition 2.} Proposition 2 guarantees that, in expectation and with high probability, the top-ranked predicted parent of a query node will be near its true parent in the knowledge graph \textit{even when the prediction is incorrect}.
Hence, if the top-ranked predicted parent of a query node is incorrect, humans-in-the-loop need only probe a small (in expectation) graph neighborhood around the top-ranked predicted parent to find the true parent of the query node. We expect this probing to require less effort than searching the entire knowledge graph.

Further, our human friendliness guarantee does not compromise ranking quality, and holds for any score function $s(u,v|\Theta)$. Hence, future research can improve the score function and the loss function minimization technique to improve ranking quality, while retaining our human friendliness guarantee without additional effort.

A limitation of Proposition 2 is its reliance on independent and identically distributed data (which is a common assumption in machine learning theory). We delegate derivations of theoretical guarantees after relaxing this assumption to future work.

\section{Experimental Evidence on the Impact of Human-Friendliness}
\label{sec:experiment}

In Section \ref{sec:theory}, we proved that for any node to be added to the knowledge graph, its true parent and the top-ranked parent predicted by our proposed method will be ``nearby'' in expectation -- we called this property \textit{human-friendliness}. Intuitively, we expect human-friendliness to reduce the effort needed by humans-in-the-loop to find nodes' true parents when the predicted parents are inaccurate, and thus increase the performance of the human-algorithm collaboration.

However, human-friendliness could also reduce the performance of the human-algorithm collaboration by increasing humans' reliance on (or trust in) possibly-inaccurate predictions \citep{stumpf2016,wang2021explanations,ahn2021will}, or by making it harder to tell when these predictions are inaccurate. Hence, in this section, we design a controlled experiment\footnote{We pre-registered how we determined our sample size, all data exclusions, all experimental manipulations, and all experimental measures \citep{simmons201221} at: {\url{https://aspredicted.org/C7C_DXC}.}
Our experiment was approved by the Institutional Review Board at the University of Wisconsin Madison.} to test how human-friendliness impacts the performance of knowledge graph expansion with humans-in-the-loop, and to gain insight into the mechanisms by which it does so.

\subsection{Experiment Design}

\begin{figure}[t]
  \vspace{-6.25mm}
  \centering
  \includegraphics[width=1.0\textwidth]{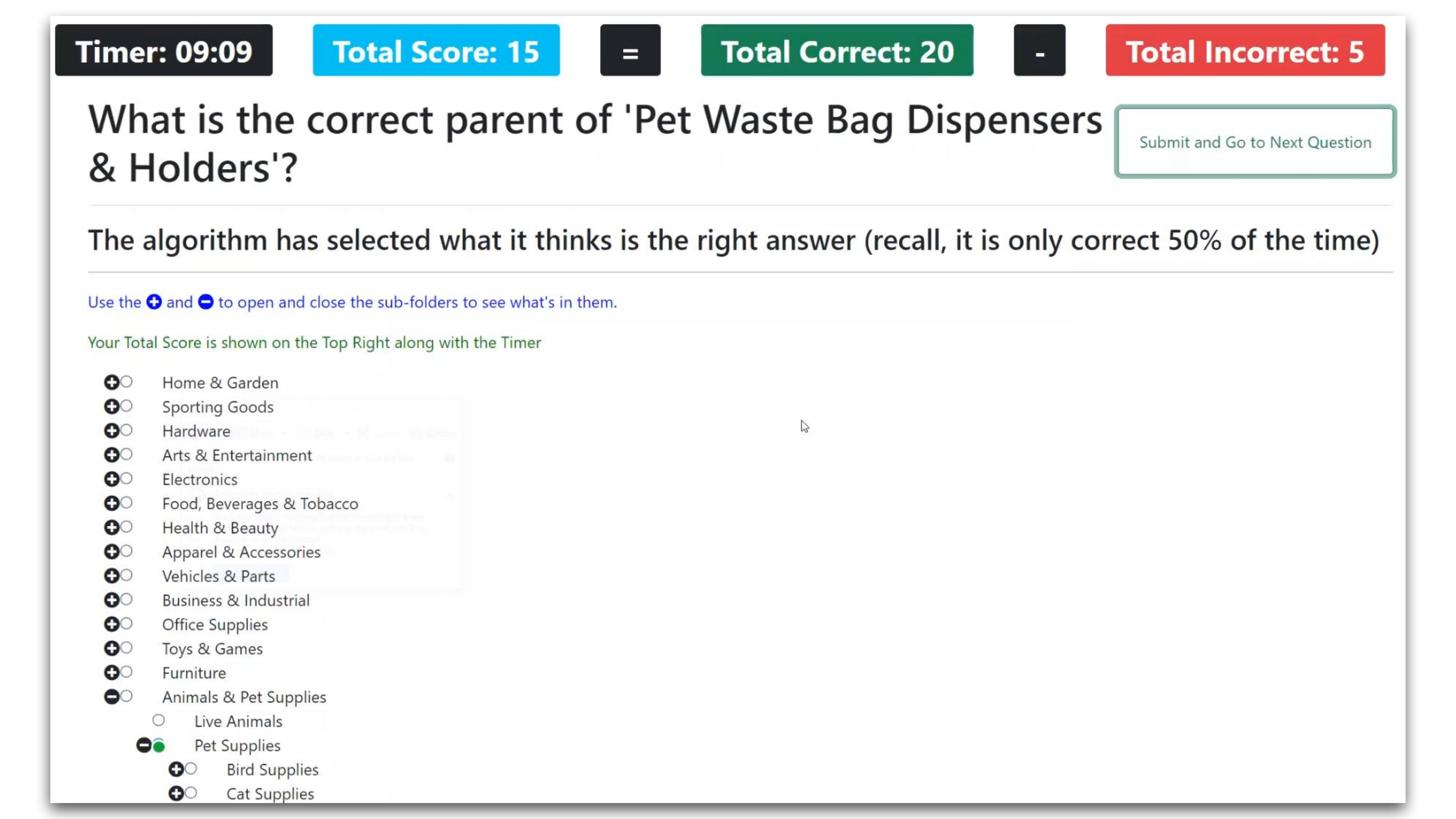}
  \caption{A screenshot of our experiment's main task with the prompt `Pet Waste Bag Dispensers \& Holders'. The time remaining and the subject's total score thus far (decomposed into the number of correct and incorrect decisions made) are displayed at the top. Levels of the hierarchy can be expanded by clicking on the `+'/`-' buttons. For each prompt, a potential parent is pre-selected (such as `Pet Supplies' in the screenshot) to support subjects' decision-making. The pre-selected parents are inaccurate for 50\% of the prompts.}
  \label{fig:exp_screenshot}
\end{figure}

\textbf{User interface description.}
In an experiment with students at a large research university in the United States, we task each subject with deciding the parents of new categories to be added to a hierarchy of product categories, with help from a decision support system. Specifically, in a web browser, we display a subset of the Google product taxonomy\footnote{The Google Product Taxonomy: \url{https://support.google.com/merchants/answer/6324436?hl=en}} and hold out 82 product categories to use as sequentially-displayed prompts (new product categories to be added to the taxonomy). The source of the product category hierarchy is not revealed to subjects. Figure \ref{fig:exp_screenshot} shows a screenshot of the user interface of our experiment's main task.

\textbf{Pre-experiment training.} Before starting the experiment, we provide subjects information about the definition of a ``parent'', information about the provided decision support, information about how their performance is scored and compensated, and hands-on practice with deciding the parents of new categories in a small product category hierarchy of beverage types. At the end of this training, each subject completes a four-prompt practice task. We record subjects' scores on the practice task for screening (detailed in our pre-registration).

During training, we inform subjects that for each prompt: (i) a potential parent is pre-selected to support their decision-making, and (ii) the pre-selected parent is incorrect 50\% of the time. We also inform subjects that they have 15 minutes for all prompts, during which they will receive 1 point for each correctly-decided parent and lose 1 point for each incorrectly-decided parent (where we consider the Google product taxonomy as ``ground truth''). Subjects receive a minimum of \$5 for participation plus a bonus of \$0.50 per point in their total score (with a minimum bonus of \$0). 

\textbf{Main experiment task.} In our experiment's main task, each subject is shown 82 sequential prompts in random order. For each prompt, subjects are asked to decide the correct parent, with decision support provided as a pre-selected potential parent; see Figure \ref{fig:exp_screenshot} for an example. In each prompt, subjects are reminded that the pre-selected potential parent is correct 50\% of the time, and can view the time remaining (out of 15 minutes) and their total score thus far (decomposed into the number of correct and incorrect decisions made) at the top of the page. The main task ends when the 15-minute timer expires or all 82 prompts have been answered.

\textbf{Treatment conditions.}
We randomly assign each subject to one of two treatment conditions with the goal of validating the impact of the human-friendliness property described at the start of this section.
Hence, our two treatment conditions systematically differ only in how ``nearby'' the pre-selected potential parents (which are provided to subjects as decision support) and the correct parents of prompts are, in terms of their undirected shortest path distance in the product category hierarchy. Specifically, our two treatment conditions are as follows:
\begin{enumerate}
    \item \textbf{``Close Errors'' or ``Human-Friendly'' (abbreviated as HF):} The pre-selected potential parents provided as decision support are correct for 50\% of the prompts. When incorrect, the pre-selected potential parent is a randomly-selected node 1-hop away from the correct parent in the product category hierarchy (with an undirected shortest path distance of 1).
    
    \item \textbf{``Far Errors'' or ``Not-Human-Friendly'' (abbreviated as NHF).}  The pre-selected potential parents provided as decision support are correct for 50\% of the prompts. When incorrect, the pre-selected potential parent is a randomly-selected node 5-hops\footnote{For each of our 82 experiment prompts, we ensure that the correct parent does indeed have nodes in the product category hierarchy that are 5-hops away in terms of the undirected shortest path distance.} away from the correct parent in the product category hierarchy (with an undirected shortest path distance of 5).
\end{enumerate}
As such, our treatment conditions mimic two decision support systems that are both equally (=50\%) accurate, but differ in their human-friendliness.

\textbf{Dependent variables.} Our pre-registered main dependent variable is subjects' total score (the total number of correct decisions minus the total number of incorrect decisions made by the end of our experiment's main task). As secondary dependent variables, we also pre-registered that we would consider the time taken by subjects to make a decision per prompt, whether their decision was correct, and whether they \textit{complied} with the decision support by adopting the pre-selected potential parent as their final decision. Finally, we pre-registered that we would investigate the impact of whether the pre-selected potential parent is correct or incorrect.

\subsection{Experimental Subjects}

We invited undergraduate and graduate students (18 years or older) at a large research university in the United States to participate through a behavioral laboratory subject pool recruitment system. We compensated subjects using Amazon.com electronic gift cards, and paid out bonuses (above the minimum payment of \$5) ranging from \$0 to \$36. Our analysis includes 54 subjects in the Human-Friendly (HF) condition and 50 subjects in the Not-Human-Friendly (NHF) condition, after applying our pre-registered rules for sample size and exclusions.

\begin{figure}[t]
  \centering
  \begin{subfigure}[b]{0.245\textwidth}
    \centering
    \includegraphics[width=\linewidth]{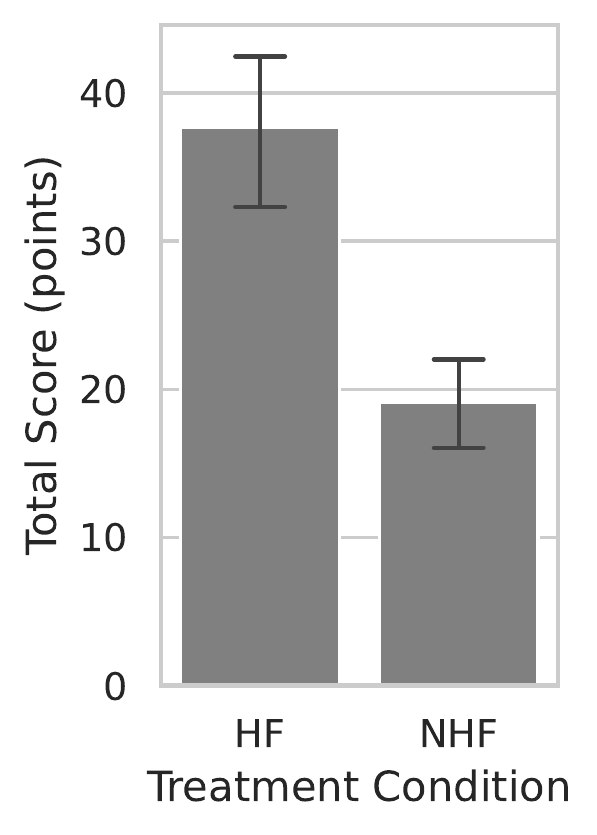}
    \caption{Total Score}
  \end{subfigure}
  \begin{subfigure}[b]{0.245\textwidth}
    \centering
    \includegraphics[width=\linewidth]{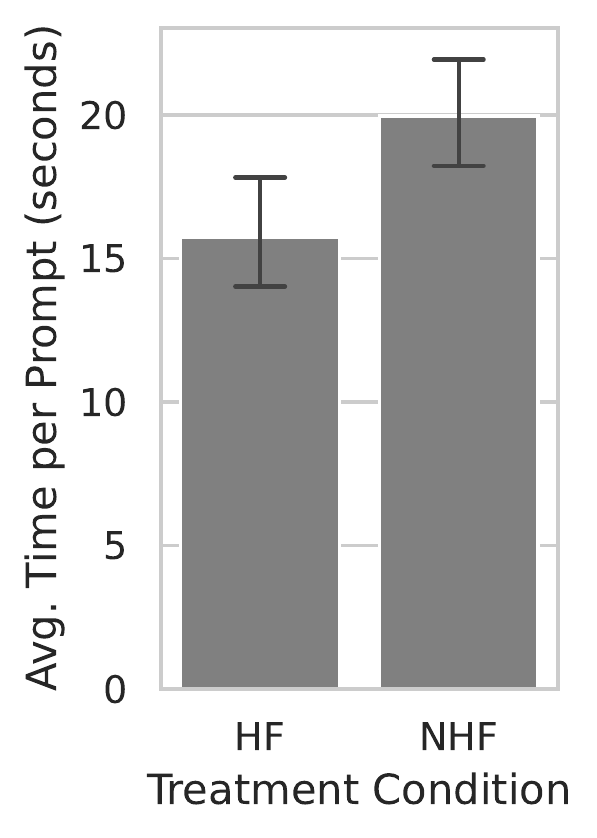}
    \caption{Time per Prompt}
  \end{subfigure}
  \begin{subfigure}[b]{0.245\textwidth}
    \centering
    \includegraphics[width=\linewidth]{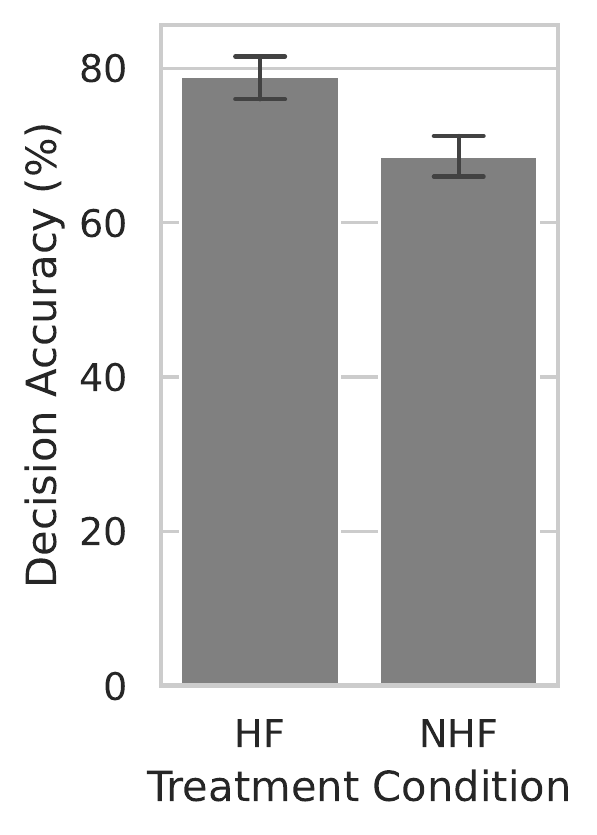}
    \caption{Decision Accuracy}
  \end{subfigure}
  \begin{subfigure}[b]{0.245\textwidth}
    \centering
    \includegraphics[width=\linewidth]{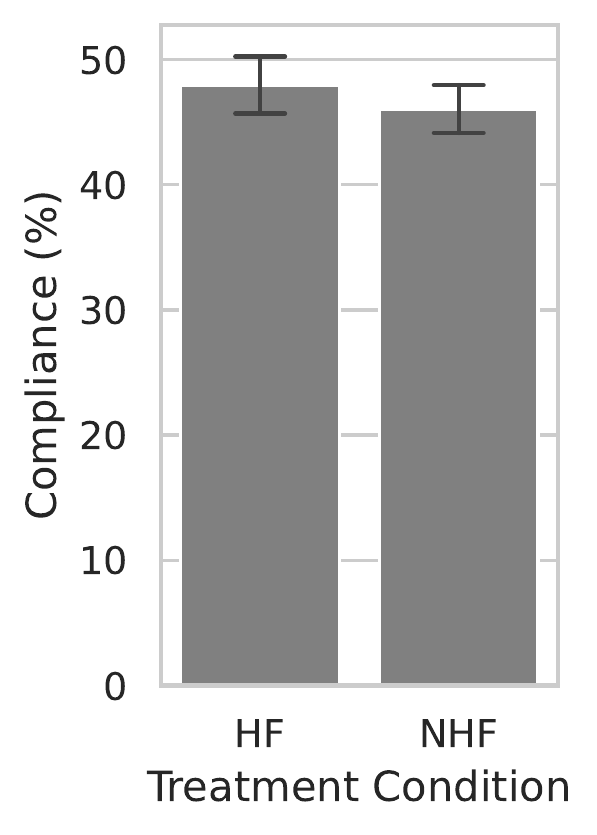}
    \caption{Compliance}
  \end{subfigure}
  \caption{The bar plots above show (a) the total score, (b) the average decision time per prompt, (c) the percentage of decisions that were correct (decision accuracy), and (d) the percentage of decisions that followed the pre-selected potential parent (compliance with the decision support), for subjects in the Human-Friendly (HF) and Not-Human-Friendly (NHF) treatment conditions. Error bars denote standard errors. \label{fig:expresults}}
\end{figure}
\subsection{The Impact of Human-Friendliness on Total Scores}
\label{subsubsection:totalscoreresults}

In Figure \ref{fig:expresults}(a), we compare the total score for subjects in the Human-Friendly (HF) and Not-Human-Friendly (NHF) treatment conditions. We find that subjects in the Human-Friendly (HF) condition score 97\% more points on average than subjects in the Not-Human-Friendly (NHF) condition: subjects in the Human-Friendly (HF) condition have an average total score of 37.7 points (s.d. = 19.1, median = 35.5), while subjects in the Not-Human-Friendly (NHF) condition have an average total score of 19.1 points (s.d. = 11.7, median = 19.0). The difference in the average total scores between the two conditions is statistically significant based on our pre-registered one-sided Mann-Whitney U-test  ($p$=9.1$\times$10$^{-8}$, $U$=2152), and on a one-sided Welch's t-test ($p$=3.7$\times$10$^{-8}$, $t$=5.9).

\begin{figure}[t]
  \centering
  \begin{subfigure}[t]{0.245\textwidth}
    \centering
    \includegraphics[width=1.19\linewidth]{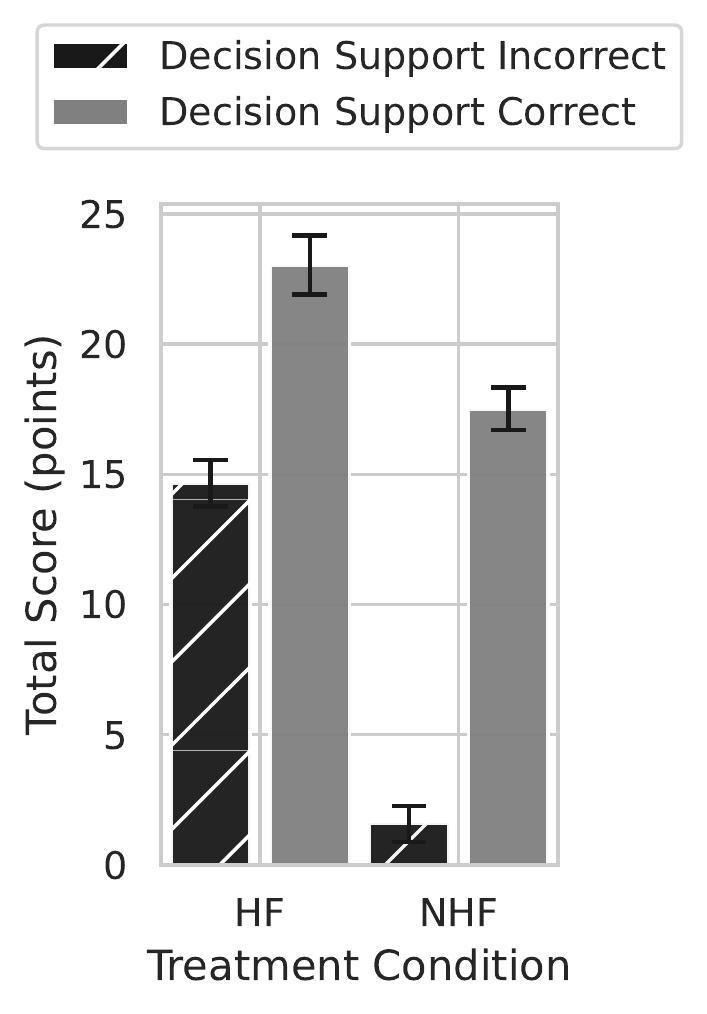}
    \caption{Total Score}
  \end{subfigure}
  \begin{subfigure}[t]{0.245\textwidth}
    \centering
    \includegraphics[width=\linewidth]{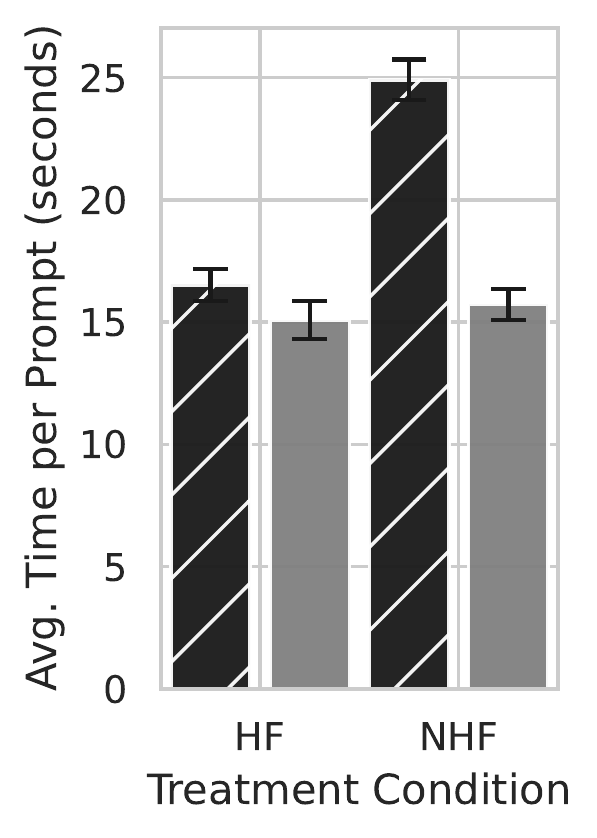}
    \caption{Time per Prompt}
  \end{subfigure}
  \begin{subfigure}[t]{0.245\textwidth}
    \centering
    \includegraphics[width=\linewidth]{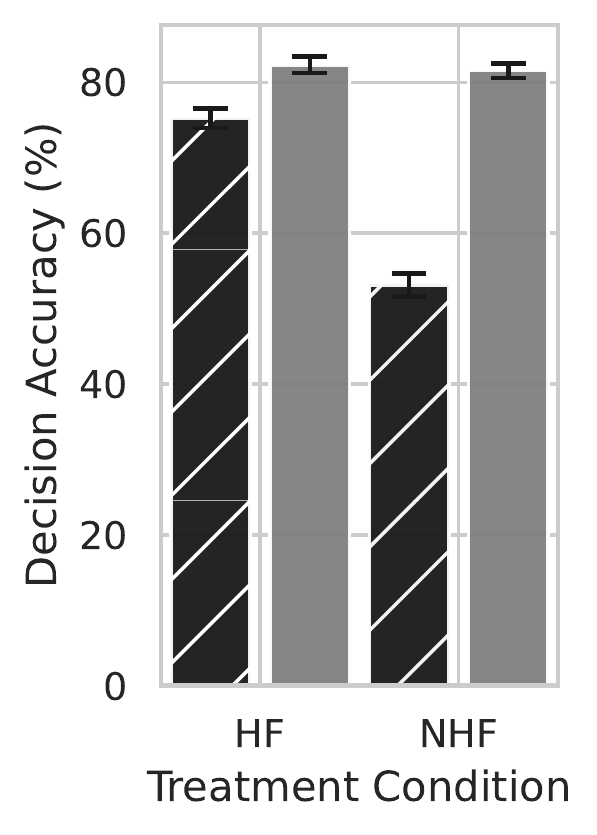}
    \caption{Decision Accuracy}
  \end{subfigure}
  \begin{subfigure}[t]{0.245\textwidth}
    \centering
    \includegraphics[width=\linewidth]{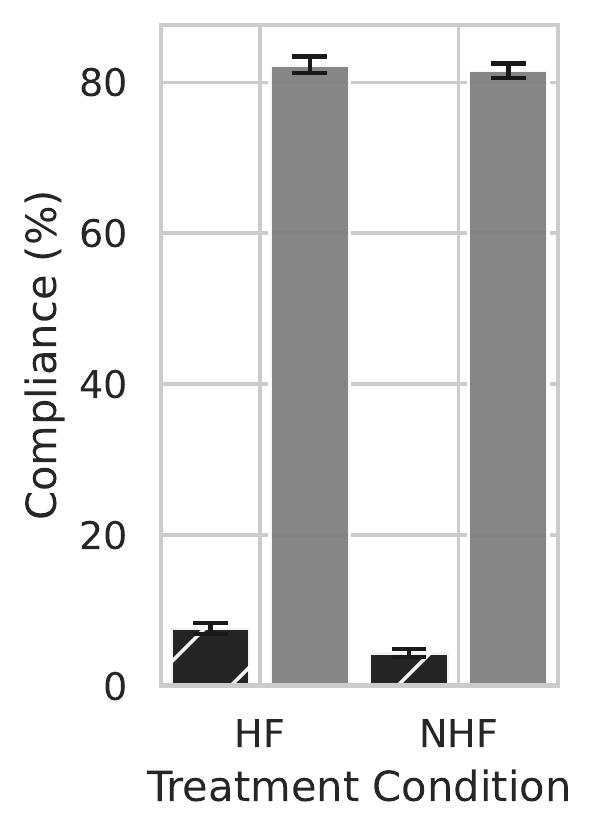}
    \caption{Compliance}
  \end{subfigure}
  \caption{The bar plots above show the 4 outcomes from Figure \ref{fig:expresults} conditional on prompts where the decision support (pre-selected potential parent) was incorrect and correct, for subjects in the Human-Friendly (HF) and Not-Human-Friendly (NHF) treatment conditions. Error bars denote standard errors. \label{fig:expresultsbyalg}}
\end{figure}

\textbf{Heterogeneity with decision support correctness.} In Figure \ref{fig:expresultsbyalg}(a), we further decompose the impact of human-friendliness on total scores by separately considering prompts where the decision support is incorrect and when it is correct. We find that human-friendliness improves total scores both when the decision support is correct and when it is incorrect, and that the magnitude of improvement is greater when the decision support is incorrect.

Specifically, conditional on prompts where the decision support is correct, subjects in the Human-Friendly (HF) condition have a total score of 23.0 points on average (s.d. = 11.6), while subjects in the Not-Human-Friendly (NHF) condition have a total score of 17.5 points on average (s.d. = 8.3). Conditional on prompts where the decision support is incorrect, subjects in the Human-Friendly (HF) condition have a total score of 14.7 points on average (s.d. = 9.2), while subjects in the Not-Human-Friendly (NHF) condition have a total score of 1.6 points on average (s.d. = 7.1).

In Appendix A, we assess the statistical significance of the aforementioned differences in the average total scores by treatment condition after conditioning on the decision support correctness (see column 1 in Tables \ref{tab:expreg} and \ref{tab:expAMEs}). We find that all differences are statistically significant ($p$ < 0.01).

\subsection{The Impact of Human-friendliness on Speed and Accuracy}

We attribute the impact of human-friendliness on subjects' total scores to improvements in their speed and accuracy. Hence, in Figure \ref{fig:expresults}(b) and \ref{fig:expresults}(c), we compare the average decision time per prompt and the decision accuracy (respectively) for subjects in the Human-Friendly (HF) and Not-Human-Friendly (NHF) conditions.

We find that subjects in the Human-Friendly (HF) condition make decisions faster (mean = 15.8 seconds, s.d. = 7.0) than subjects in the Not-Human-Friendly (NHF) condition (mean = 20.0 seconds, s.d. = 7.1). The difference between these means is statistically significant based on a one-sided Mann-Whitney U-test  ($p$=2.3$\times$10$^{-4}$, $U$=812), and on a one-sided Welch's t-test ($p$=1.6$\times$10$^{-3}$, $t$=-3.0).

We also find that subjects in the Human-Friendly (HF) condition make decisions more accurately (mean = 78.9\%, s.d. = 10.6\%) than subjects in the Not-Human-Friendly (NHF) condition (mean = 68.6\%, s.d. = 10.1), despite the decision support being 50\% accurate in both conditions. The difference between these means is statistically significant based on a one-sided Mann-Whitney U-test  ($p$=5.6$\times$10$^{-7}$, $U$=2099), and on a one-sided Welch's t-test ($p$=9.2$\times$10$^{-7}$, $t$=5.1).

\textbf{Heterogeneity with decision support correctness.} In Figure \ref{fig:expresultsbyalg}(b) and \ref{fig:expresultsbyalg}(c), we further decompose the impact of human-friendliness on speed and accuracy by separately considering prompts where the decision support is incorrect and when it is correct. We find that human-friendliness results in a statistically significant ($p < 0.01$) decrease in the average decision time per question and a statistically significant ($p < 0.01$) increase in the decision accuracy when the decision support is incorrect, but not when it is correct.

Specifically, conditional on prompts where the decision support is incorrect, subjects in the Human-Friendly (HF) condition have an average decision time per prompt of 16.5 seconds (s.d. = 6.7), while subjects in the Not-Human-Friendly (NHF) condition have an average decision time per prompt of 24.9 seconds (s.d. = 8.5). Conditional on prompts where the decision support is correct, subjects in the Human-Friendly (HF) condition have an average decision time per prompt of 15.1 seconds (s.d. = 7.9), while subjects in the Not-Human-Friendly (NHF) condition have an average decision time per prompt of 15.7 seconds (s.d. = 6.5).

In addition, conditional on prompts where the decision support is incorrect, subjects in the Human-Friendly (HF) condition have an accuracy of 75.2\% (s.d. = 13.4\%), while subjects in the Not-Human-Friendly (NHF) condition have an accuracy of 53.1\% (s.d. = 15.4\%). Conditional on prompts where the decision support is correct, subjects in the Human-Friendly (HF) condition have an accuracy of 82.3\% (s.d. = 11.3\%), while subjects in the Not-Human-Friendly (NHF) condition have an accuracy of 81.6\% (s.d. = 9.6\%).

In Appendix A, we assess the statistical significance of the aforementioned differences in the speed and accuracy by treatment condition after conditioning on the decision support correctness (see columns 2 and 3 in Tables \ref{tab:expreg} and \ref{tab:expAMEs}).

\subsection{The Impact of Human-Friendliness on Compliance with the Decision Support}

In Figure \ref{fig:expresults}(d), we compare the percentage of prompts for which subjects make decisions identical to the decision support (the pre-selected potential parent), for subjects in the Human-Friendly (HF) and Not-Human-Friendly (NHF) conditions. We find that subjects in the Human-Friendly (HF) condition follow the decision support more (mean = 48.0\%, s.d=7.8\%) than subjects in the Not-Human-Friendly (NHF) condition (mean = 46.0\%, s.d. = 7.0\%). However, the difference between these means is not statistically significant based on a one-sided Mann-Whitney U-test  ($p$=0.06, $U$=1590), or on a one-sided Welch's t-test ($p$=0.09, $t$=1.33).

\clearpage

\textbf{Heterogeneity with decision support correctness.} In
Figure \ref{fig:expresultsbyalg}(d), we further decompose the impact of human-friendliness on compliance with the decision support by separately considering prompts where the decision support is incorrect and when it is correct. We find that human-friendliness leads to a statistically significant ($p < 0.05$) increase in compliance with the decision support when the decision support is incorrect, but not when it is correct.

Specifically, conditional on prompts where the decision support is incorrect, subjects in the Human-Friendly (HF) condition follow the decision support for 7.6\% of the prompts on average (s.d. = 7.2\%), while subjects in the Not-Human-Friendly (NHF) condition follow the decision support for 4.3\% of the prompts on average (s.d. = 5.3\%). Conditional on prompts where the decision support is correct, subjects in the Human-Friendly (HF) condition follow the decision support for 82.3\% of the prompts on average (s.d. = 11.3\%), while subjects in the Not-Human-Friendly (NHF) condition follow the decision support for 81.6\% of the prompts on average (s.d. = 9.6\%).

In Appendix A, we assess the statistical significance of the aforementioned differences in compliance with the decision support by treatment condition after conditioning on the decision support correctness (see column 4 in Tables \ref{tab:expreg} and \ref{tab:expAMEs}).

\subsection{Summary and Discussion}
\label{sec:experimentsummary}

With a controlled experiment, we provide evidence that our human-friendliness property improves the performance of knowledge graph expansion with humans-in-the-loop. Despite the provided decision support being 50\% accurate in both  treatment conditions, subjects assigned to the Human-Friendly (HF) treatment condition scored 97\% more on our knowledge graph expansion task than subjects assigned to the Not-Human-Friendly (NHF) treatment condition. 

We further generate insight into the mechanism by which human-friendliness improves performance, and find that human-friendliness increased both humans' decision-making speed and accuracy (by about 21\% and 15\%, respectively). Notably, the impact of human-friendliness is particularly large in instances when the decision support is incorrect. In these instances, subjects assigned to the Human-Friendly (HF) condition scored 800\% more, were 34\% faster, and were 42\% more accurate than subjects assigned to the Not-Human-Friendly (NHF) condition. This suggests that human-friendliness improves human-in-the-loop performance primarily by increasing humans' speed and accuracy in fixing the errors of the provided decision support.

Finally, we evaluate the potential negative impact of human-friendliness on performance by increasing humans' reliance on the possibly-inaccurate decision support. We indeed find weak evidence that human-friendliness increases humans' compliance with the decision support even when it is incorrect.
However, in our experiment, this concern is far outweighed by the benefits of human-friendliness. Conditional on the decision support being incorrect, human-friendliness still improves accuracy because when subjects do attempt to correct the decision support, they do it faster and more accurately.

\section{Case Study on the Pinterest Knowledge Graph}
\label{sec:casestudy}

The Pinterest knowledge graph is a network of phrases called ``interests''.
These interests are used to tag users, search queries, and other Pinterest content \citep{pinterest2020taxonomy}. Interests subsequently serve as the targets of contextual ads \citep{pinterest2022interesttargeting}. In addition, the network structure of the knowledge graph is used to ``expand'' interests, broaden targeted audience segments, and improve ads' reach \citep{pinterest2019expanded}.

Due to its central role in Pinterest's advertising and recommendation pipeline, it is critical to maintain the quality and completeness of the Pinterest knowledge graph. Hence, in this section, we use the Pinterest knowledge graph to benchmark the accuracy and human-friendliness of our proposed knowledge graph expansion method against several baselines. Our proposed method was successfully deployed at Pinterest to help knowledge graph curators in the most recent knowledge graph expansion effort; we discuss this deployment in Section \ref{sec:deployment}.

\subsection{Data}
\label{sec:data}

The Pinterest knowledge graph consists of 24 directed subgraphs with 10,791 nodes and 10,768 edges (additional graph statistics are in Table \ref{tab:data}). The subgraphs are hierarchical, with each edge directed from a more-specific child node to a more-general parent node (such as ``Oral Care'' $\rightarrow$ ``Health''). The most general node of each subgraph is called its ``root''. Each non-root node has exactly one parent. To ensure that a path exists between every pair of nodes in the knowledge graph, we add a dummy knowledge graph root linked to each subgraph root as its parent.
%

\clearpage

\textbf{Constructing training, validation, and test datasets.}
We hold out a random sample of 30\% of the leaf nodes (which have no children) for testing. Of the remaining nodes, we randomly sample 85\% for training and use the rest for validation. The resulting training, validation, and test datasets are lists of node pairs $(u, v)$, where $u$ is the child node and $v$ is its parent. Each child node $u$ belongs to exactly one of the training, validation, or test node sets. Holding out leaf nodes for testing ensures that a test child node is never previously seen as a non-test parent node. Since the subgraph roots have no parents, we exclude them from appearing as child nodes in our datasets.

\textbf{Sampling negative training pairs.} For training, all the methods we benchmark require both ``positive'' pairs (child-parent node pairs) and ``negative'' pairs (node pairs from the graph that are \textit{not} child-parent node pairs). However, including all possible negative pairs when training is computationally infeasible\footnote{There are $\sim$8,000 negative pairs per training child node in the graph ($\sim$63 million negative pairs in total). When training any of the benchmarked methods with all negative pairs, we exceed the 512GB of memory we have available on our development servers. Training is feasible after sampling 5,000 negative pairs ($\sim$60\%) per training child node.}. Hence, for each training child node, we randomly sample 5,000 training nodes excluding its parent to construct negative training pairs. Sampling negative training pairs is standard in the information retrieval and ranking literature \citep{steffen2009bpr}.

\textbf{Generating input features.} Our proposed method and several baselines require input features for each node of the knowledge graph. Hence, we use the word embedding of the text of each node as its input features. We generate 300-dimensional embeddings using two approaches: (i) FastText \citep{bojanowski2017enriching} using the crawl-300d-2M-subword pretrained model \citep{mikolov2018advances}, and (ii) PinText \citep{zhuang2019pintext} using a model trained in-house at Pinterest.

The PinText embeddings are trained on millions of tokens from the Pinterest platform\footnote{The PinText embeddings we use were generated before the Pinterest knowledge graph was first deployed at Pinterest, and are hence not contaminated with information from the Pinterest knowledge graph.}, whereas the FastText embeddings are trained on 6.3 billion tokens from the Common Crawl website archive.
Thus, the two embedding approaches have different strengths and weaknesses: the PinText embeddings are more relevant to our context but less representative of language, whereas the FastText embeddings are less relevant to our context but more representative of language.

For both embedding approaches, we use mean-pooling \citep{shen2018baseline} (averaging the embeddings of the constituent words of a phrase) to generate the embeddings of multi-word phrases, and subword information to handle out-of-vocabulary words \citep{bojanowski2017enriching}. We then normalize the embeddings to have unit $L_2$ norm based on prior work \citep{mikolov2013distributed}.

\clearpage

\begin{table}[t]
  \vspace{-7mm}
  \begin{tabular}{l@{\hskip 0.5in}l}
    \toprule
    Number of nodes (excluding the dummy root) & 10791 \\
    \quad Number of training child nodes (= number of training node pairs) & 6711 \\
    \quad Number of validation child nodes (= number of validation node pairs) & 1183 \\
    \quad Number of test child nodes (= number of test node pairs) & 2872 \\
    Number of leaf nodes & 9576 \\
    Number of edges (excluding edges between the dummy root and subgraph roots) & 10768 \\
    Maximum undirected shortest path distance (diameter) & 16 \\
    Mean undirected shortest path distance & 7.74 \\
    \bottomrule
  \end{tabular}
  \caption{\textbf{Graph statistics}. The Pinterest knowledge graph consists of 24 directed hierarchical subgraphs.}
  \label{tab:data}
\end{table}


\subsection{Evaluation Metrics}
\label{sec:metrics}

We evaluate each method on a test set of nodes $\{u_1, \dots, u_M\}$ that are held out from the knowledge graph and never seen during training, as described in Section \ref{sec:data}.
For each test node $u_i$, let $v_i$ be its true parent, $L(u_i)$ be the ranked list of predicted parents (which contains all non-test nodes) produced by the method being evaluated, and rank$_i$ be the position of the true parent $v_i$ in $L(u_i)$.

\textbf{Ranking quality.} We quantify the ranking quality of each method using the following two metrics that are standard in the evaluation of ranking methods \citep{lin2021pretrained}:
\begin{enumerate}
  \item \textbf{Mean Reciprocal Rank (MRR):} The mean reciprocal rank is the average of the reciprocal ranks of the true parents of the test nodes, given by:
  \begin{align}
    \textrm{MRR} = \frac{1}{M} \sum_{i=1}^M \frac{1}{\textrm{rank}_i} \times 100
  \end{align}
  The MRR is 100\% (best) when the true parent of every test node is ranked first in the list of predicted parents. The MRR tends to 0\% (worst) as the true parents of test nodes are ranked lower in the lists of predicted parents.

  \item\textbf{Recall at 1 (R@1):} The recall at 1 is the fraction of test nodes for which the top-ranked predicted parent is the true parent:
  \begin{align}
    \textrm{R@1} = \frac{1}{M} \sum_{i=1}^M \mathbb{I}[\textrm{rank}_i = 1] \times 100
  \end{align}
  where $\mathbb{I}[\cdot]$ is the indicator function. The R@1 is 100\% (best) when the true parent of every test node is the top-ranked predicted parent. If none of the test nodes have their true parents identical to the top-ranked predicted parents, the R@1 is 0\% (worst).
\end{enumerate}

\textbf{Human-friendliness.} We quantify human-friendliness by measuring how ``nearby'' in the graph the true parents and top-ranked predicted parents of each test node are on average. While the undirected shortest path distance is an intuitive measure of ``nearness'', it ranges from 0 (best) to the graph diameter (worst). Hence, we normalize the undirected shortest path distance by the graph diameter (the maximum undirected shortest path distance between any node pair).

For each test node $u_i$, let $\hat{v}(u_i)$ be its top-ranked predicted parent and $d_{\mathcal{G}}(v_i, \hat{v}(u_i))$ be the undirected shortest path distance between its true parent $v_i$ and its top-ranked predicted parent $\hat{v}(u_i)$ in the training knowledge graph $\mathcal{G}$. Further, let $D$ be the diameter of $\mathcal{G}$. We quantify human-friendliness using the following metrics:
\begin{enumerate}
  \item \textbf{Mean Normalized Shortest Path Distance (MND):} The mean normalized shortest path distance is the average of the undirected shortest path distances between the true parents and top-ranked predicted parents of each test node, normalized by the graph diameter:
  \begin{align}
    \textrm{MND} = \frac{1}{M} \sum_{i=1}^M \frac{d_{\mathcal{G}}(v_i, \hat{v}(u_i)))}{D} \times 100
  \end{align}
  The MND is 0\% (best) when the top-ranked predicted parent is the true parent for every test node. When the top-ranked predicted parents are at a distance $D$ from the true parents of every test node, the MND is 100\% (worst).
  \item \textbf{Mean Normalized Shortest Path Distance when Incorrect (MND-I):}
  The MND is correlated with the MRR and R@1, since $d_{\mathcal{G}}(v_i, \hat{v}(u_i)) = 0$ when the top-ranked predicted parent $\hat{v}(u_i)$ is the true parent $v_i$. Hence, we also compute the mean normalized shortest path distance only considering those test nodes for which the top-ranked predicted parent is \textit{not} the true parent:
  \begin{align}
    \textrm{MND-I} = \frac{1}{\sum_{j=1}^M \mathbb{I}[v_j \neq \hat{v}(u_j)]} \sum_{i=1}^M \frac{d_{\mathcal{G}}(v_i, \hat{v}(u_i))}{D} \times \mathbb{I}[v_i \neq \hat{v}(u_i)] \times 100
  \end{align}
  where $\mathbb{I}[\cdot]$ is the indicator function. The MND-I metric quantifies how human-friendly a method is when it is incorrect. As such, the MND-I captures human-friendliness ``above and beyond'' the recall at 1 (R@1). The MND-I ranges between 0\% (best) and 100\% (worst). If the top-ranked predicted parent is always correct, the MND-I is undefined.
\end{enumerate}

\subsection{Designing the Score Function}
\label{sec:relatedness}

We design a score function for our case study based on two key insights acquired via conversations with knowledge graph curators at Pinterest. First, edges in the Pinterest knowledge graph have unobserved and heterogeneous semantic meanings. For example, edges between city and country nodes (which have \textsc{Is-In} semantics) coexist with edges between clothing categories (which have \textsc{Is-Type-Of} semantics) without being explicitly labeled. Second, all the incoming edges at a given node have identical (but unobserved) semantic meanings. For example, all the edges into the node ``France'' from its children have \textsc{Is-In} semantics.

We begin with a simple score function parameterized by $\Theta = \{M\}$ for each pair of nodes $u$ and $v$ having feature vectors $\pmb{e}_u \in \mathbb{R}^d$ and $\pmb{e}_v \in \mathbb{R}^d$ respectively:
\begin{align}
  s(u,v|\Theta) = (\mathbf{e}_u M) \cdot \mathbf{e}_v
  \label{eq:crim}
\end{align}
Intuitively, $M \in \mathbb{R}^{d\times d}$ represents each knowledge graph edge as a linear transformation in the nodes' feature vector space. However, a single linear transformation for all nodes may not have sufficient representational capacity to capture heterogeneous edge semantics. Hence, we consider replacing $M$ with a different linear transformation $M_v \in \mathbb{R}^{d\times d}$ for each node $v$:
\begin{eqnarray}
  s(u, v|\Theta) = (\mathbf{e}_u M_v) \cdot \mathbf{e}_v
  \label{eq:similarity}
\end{eqnarray}
While this score function has a larger representational capacity, it greatly increases the number of parameters to learn (especially in large knowledge graphs). Specifically, if the training knowledge graph has $|V|$ nodes, the number of parameters to learn is $O(d^2|V|)$. Additionally, this score function fragments the training knowledge graph into node-level subsets, with the linear transformation $M_v$ at each node $v$ being learned only from that node's children. As such, there is limited information-sharing between nodes during training.

\clearpage

~\\[-11mm]
\indent Hence, we decompose each linear transformation $M_v$ into $k$ linear transformations $P_1 \in \mathbb{R}^{d\times d}, \dots, P_k \in \mathbb{R}^{d\times d}$ that are shared among all knowledge graph nodes:
\begin{eqnarray}
  M_v = \sum_{i=1}^k \mathbf{w}_v[i] \times P_i
\end{eqnarray}
where $\mathbf{w}_v \in \mathbb{R}^k$ is a weight vector associated with each node to be learned from the training knowledge graph, and $k$ is a hyperparameter. This score function allows different nodes to share information with each other through $P_1, \dots, P_k$. Intuitively, $P_1, \dots, P_k$ capture $k$ ``global'' edge semantics, and each edge is represented as a linear combination of these global semantics.

To further facilitate information-sharing between nodes and to reduce the number of parameters to learn, we define the weight vectors in terms of the node feature vectors as follows:
\begin{eqnarray}
  \mathbf{w}_v = f(\mathbf{e_v})
\end{eqnarray}
where $f: \mathbb{R}^d \rightarrow \mathbb{R}^k$ is any learnable function (we use a feedforward neural network with Tanh activations). Our final score function is parameterized by $\Theta = \{k, P_1, \dots, P_k, f\}$, and the number of parameters to learn is $O(|f| + d^2k)$, which is independent of the size of the knowledge graph $|V|$.

\subsection{Baselines}
\label{sec:baselines}

We compare our proposed method with two naive baselines and several state-of-the-art methods from the machine learning literature\footnote{We discuss hyperparameter tuning and implementation details for all benchmarked methods in Appendix B.}. As our first naive baseline, we randomly permute the training nodes to generate the ranked list of parents for each test child node; we call this the Random Guess baseline. As our second naive baseline, we score node pairs based on the Jaccard similarity\footnote{The Jaccard similarity of a text pair is the number of common words divided by the total number of unique words in the text pair. For example, the Jaccard similarity of ``Modern Design'' and ``Design and Architecture'' is 1/4.} of their raw texts. The naive baselines do not use our training data or input features.

\textbf{Feedforward neural network (FFNN).} As our next baseline, we use a feedforward neural network trained as a binary classifier (minimizing a cross-entropy loss) to distinguish between the positive and negative training node pairs. We use the output of the final sigmoid layer to score node pairs and produce a ranking of possible parents for each test child node. We use the concatenated word embeddings of the text of each pair of nodes as input features to the neural network.

\textbf{LambdaMART.} We then benchmark our proposed method against LambdaMART \citep{burges2010from}. LambdaMART is an ensemble of gradient boosted decision trees trained to rank positive training node pairs higher than negative training node pairs. It is the state-of-the-art in ranking structured data, outperforming neural network-based methods on recent benchmarks \citep{qin2020neural}. We use the LightGBM \citep{ke2017lightgbm} implementation of LambdaMART, and use the concatenated word embeddings of the text of each pair of nodes as input features to LambdaMART.

\textbf{Fine-tuned transformers.} In ranking unstructured text data, pretrained transformers \citep{vaswani2017attention} fine-tuned for ranking are the current state-of-the-art \citep{lin2021pretrained}.
Fine-tuning involves further training a pretrained transformer as a binary classifier to distinguish between positive and negative training pairs. After fine-tuning, transformers produce a score for each node pair, which is then used to produce a ranking of possible parents for each test child node. In contrast with the other benchmarked methods, transformers operate on raw text and not on input features.

We select pretrained transformers to fine-tune based on their performance on text ranking benchmarks\footnote{\url{https://www.sbert.net/docs/pretrained-models/ce-msmarco.html}}. We first fine-tune the MiniLM-L12-H384-uncased pretrained transformer \citep{wang2020minilm} on the text of our training node pairs. We then fine-tune two variants of MiniLM-L12-H384-uncased that were pretrained on additional external data (note that this additional data is not made available to other benchmarked methods):
\begin{enumerate}
  \item all-MiniLM-L6-v2\footnote{\url{https://huggingface.co/sentence-transformers/all-MiniLM-L6-v2}}: This transformer is additionally pretrained on 1.2 billion related text pairs from various online sources (such as question-answer pairs from Yahoo Answers and duplicate question pairs from StackExchange) to distinguish them from unrelated text pairs.
  
  \item ms-marco-MiniLM-L6-v2\footnote{\url{https://huggingface.co/cross-encoder/ms-marco-MiniLM-L-6-v2}}: This transformer is additionally pretrained on the 1.2 billion text pairs used to pretrain (i) and on 0.5 million query-document pairs from the MS-MARCO passage ranking dataset \citep{nguyen2016ms} with the goal of accurately ranking documents. This produces state-of-the-art results on the MS-MARCO passage ranking benchmark$^4$.
\end{enumerate}
We obtain the pretrained transformer weights from the Hugging Face Hub \citep{huggingface} and fine-tune them using the sentence-transformers library \citep{reimers2019sentence}.

\textbf{Triplet matching network (TMN).} Finally, we benchmark our proposed method against the triplet matching network \citep{zhang2021taxonomy}, which is a state-of-the-art knowledge graph expansion method. The triplet matching network is trained to predict both the parent and the child of nodes in the training data, and can hence also be used to insert non-leaf nodes into the graph. We use the authors' public implementation of this method\footnote{\url{https://github.com/JieyuZ2/TMN}}, and use the concatenated word embeddings of the text of each pair of nodes as input features to the triplet matching network.


\subsection{Benchmarking Results}
\label{sec:results}

\begin{wrapfigure}[8]{r}{0.35\textwidth}
  \vspace{-1in}
  \begin{center}
    \includegraphics[width=\textwidth]{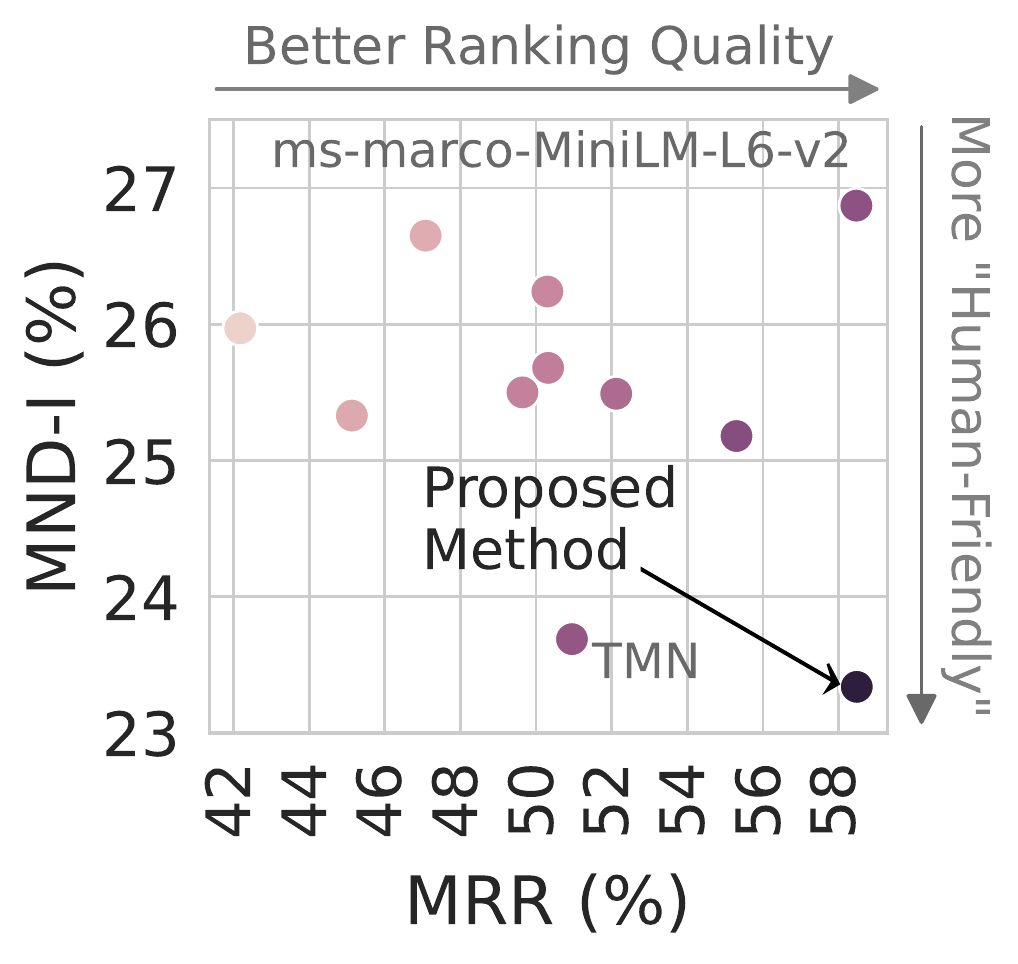}
  \end{center}
  \vspace{-0.3in}
  \caption{Benchmarked methods compared on ranking quality (MRR) and human-friendliness (MND-I); see discussion in Section \ref{sec:results} for details.}
  \label{fig:resultssummary}
\end{wrapfigure}
Table \ref{tab:mainresults} shows our benchmarking results. Figure \ref{fig:resultssummary} summarizes the key takeaway from our results by plotting each method (excluding the naive baselines) along two dimensions: (i) ranking quality as quantified by a higher MRR, and (ii) human-friendliness as quantified by a lower MND-I. Our proposed method is the only benchmarked method to perform well on both dimensions. While the triplet matching network (TMN) is comparable on human-friendliness, it is worse on ranking quality. Similarly, the ms-marco-MiniLM-L6-v2 fine-tuned transformer is comparable on ranking quality but worse on human-friendliness. In the rest of this section, we discuss additional takeaways from the results in Table \ref{tab:mainresults} with a focus on the MRR and MND-I metrics.

\textbf{Naive baselines.} Random guessing results in a poor MRR, but achieves an MND-I of 42.58\%. This implies that a randomly and incorrectly guessed parent is \textit{less} than half the diameter of the graph away from the true parent of a test child node on average. We attribute this to the fact that the average shortest path distance in a tree-structured graph can be less than half of its diameter \citep{parhami2013exact}. The Jaccard similarity baseline improves over random guessing, but is still 60-70\% worse than other benchmarked methods on the MRR and 40-60\% worse on the MND-I.

\textbf{Feedforward neural network (FFNN).} With PinText embeddings as input features, the feedforward neural network is among the top 3 benchmarked methods on both the MRR and MND-I. This suggests that hierarchical relationships are latent in and recoverable from word embeddings, justifying our decision to use word embeddings as input features. The feedforward neural network performs relatively worse on the MRR and MND-I with FastText embeddings as input features, suggesting that the choice of word embedding technique plays a role in improving the ranking quality and human-friendliness of knowledge graph expansion.

\textbf{LambdaMART.} With PinText embeddings as input features, LambdaMART performs worse than the feedforward neural network on the MRR and MND-I. With FastText embeddings as input features, LambdaMART outperforms the feedforward neural network on the MRR but not on the MND-I. We attribute the relatively weak performance of LambdaMART to its inability to fully recover hierarchical relationships from word embeddings. In general, LambdaMART is better suited for ranking structured or tabular data and not for text \citep{qin2020neural,lin2021pretrained}.

\textbf{Fine-tuned transformers.}
The MRR of the fine-tuned transformers improves with additional pretraining from 42.18\% for MiniLM-L12-H384-uncased to 58.48\% for ms-marco-MiniLM-L6-v2, which is comparable to the best among all benchmarked methods. However, \textit{this improvement in ranking quality does not translate to an improvement in human-friendliness}. With additional pretraining, the MND-I of the fine-tuned transformers deteriorates from 25.97\% for MiniLM-L12-H384-uncased to 26.87\% for ms-marco-MiniLM-L6-v2, which is the worst MND-I among all benchmarked methods excluding the naive baselines.

\textbf{Triplet matching network (TMN).} With PinText embeddings as input features, the triplet matching network achieves an MND-I of 23.69, which is comparable to the best among all benchmarked methods. However, it achieves an MRR of 50.95\%, which is among the bottom 3 of all benchmarked methods excluding the naive baselines. The triplet matching network performs relatively worse with FastText embeddings as input features on both the MRR and MND-I.

\subsection{Discussion and Examples}

The weak association between the MRR and MND-I in Table \ref{tab:mainresults} highlights the need for theoretical guarantees. Due to the theoretical guarantees we provide in Section \ref{sec:theory}, improving our proposed method (with a better score function or optimization procedure, for example) will improve both its ranking quality and human-friendliness. In contrast, improving the other benchmarked methods that lack theoretical guarantees may lead to an unexpected tradeoff between improving the ranking quality and improving human-friendliness.

\clearpage

In Table \ref{tab:predictions}, we show the top 5 parents predicted by our proposed method for a sample of test child nodes from the Pinterest knowledge graph. To illustrate where our proposed method does well and where it does not, we separately show the parents predicted by our proposed method when the true parent is in the top 5 predicted parents (Table \ref{tab:predictions}, top), and when the true parent is not in the top 5 predicted parents (Table \ref{tab:predictions}, bottom).

\begin{table*}
  \begin{tabular}{l@{\hskip 1in}cccc}
    \toprule
    \textbf{Method} & \textbf{MRR (\%)} & \textbf{R@1 (\%)} & \textbf{MND (\%)} & \textbf{MND-I (\%)}\\
    \midrule
    Random Guess & 00.01 & 00.00 & 42.23 & 42.23 \\
    Jaccard Similarity & 17.58 & 13.75 & 32.66 & 37.86\\
    Feedforward Neural Network &&&&\\
    \quad with FastText embeddings & 49.64 & 37.52 & 15.93 & 25.50\\
    \quad with PinText embeddings & 52.12 & 41.07 & 15.03 & 25.50\\
    LambdaMART &&&&\\
    \quad with FastText embeddings & 50.30 & 39.23 & 15.94 & 26.24\\
    \quad with PinText embeddings & 50.32 & 39.68 & 15.49 & 25.68\\
    Fine-Tuned Transformers &&&&\\
    \quad MiniLM-L12-H384-uncased & 42.18 & 28.16 & 18.65 & 25.97 \\
    \quad all-MiniLM-L6-v2 & 47.08 & 34.63 & 17.42 & 26.65 \\
    \quad ms-marco-MiniLM-L6-v2 & \textbf{58.48} & 44.89 & 14.27 & 26.87\\
    Triplet Matching Network & & & &\\
    \quad with FastText embeddings & 45.13 & 32.41 & 15.64 & 25.33\\ 
    \quad with PinText embeddings & 50.95 & 39.58 & 14.32 & 23.69\\ 
    Proposed Method &&&&\\
    \quad with FastText embeddings & 55.30 & 42.08 & 14.58 & 25.18\\ 
    \quad with PinText embeddings & \textbf{58.48} & \textbf{45.11} & \textbf{12.81} & \textbf{23.34}\\ 
    \bottomrule
  \end{tabular}
  \caption{Performance of our proposed method and baselines on the Pinterest knowledge graph. The MRR and R@1 quantify ranking quality and range from 0\% (worst) to 100\% (best). The MND and MND-I quantify human-friendliness and range from 100\% (worst) to 0\% (best). The best results per metric are in \textbf{bold}.}
  \label{tab:mainresults}
\end{table*}
\begin{table*}
  \vspace{-4mm}
  \begin{tabular}{l@{\hskip 0.25in}l}
    \toprule
    \textbf{Test Child Node} & \textbf{Ranked List of Predicted Parents}\\
    \midrule
    \multicolumn{2}{c}{\textit{Sample of predictions with true parents in the top 5 predicted parents}}\\
    audrey hepburn
      & \textbf{actresses}, tennis players, authors, artists, baseball players\\
    singapore grand prix
      & \textbf{formula 1}, indycar series, asia travel, european travel, auto racing\\
    pilates routine
      & health, \textbf{fitness and exercises}, fitness,daily routine, work out\\
    woodcut
      &mixed media art, illustration, art, painting, \textbf{printmaking}\\
    artificial intelligence
      & science, life science, \textbf{technology}, human, digital art\\
    dark spots	& hair color, makeup color, romance movie, \textbf{skin concern}, band\\
      negroni   & rum cocktail, wine cocktails, italian cuisine, italy, \textbf{gin cocktails}\\
      \midrule
    \multicolumn{2}{c}{\textit{Sample of predictions with true parents (the phrase following $\rightarrow$) not in the top 5 predicted parents}}\\
    banksy $\rightarrow$ street art & mixed media art, digital art, art, artists, contemporary art\\
    python $\rightarrow$ reptiles & monkey, coding, band, monty python, science\\
    ironman race $\rightarrow$ running &	nascar, indycar series, auto racing, sport event, formula 1\\
    biscotti $\rightarrow$ biscuits &	italian cuisine, pasta, desserts, food and drinks, cheese\\
    ghee $\rightarrow$ butter &	chinese cuisine, food and drinks, bread, indian cuisine, appetizers\\
    flying $\rightarrow$ travel tips & aircraft, band, insects, DIY home and decorations, boats\\
    food chain $\rightarrow$ life science &	food and drinks, diet, DIY edible, pet food, frozen food\\
    \bottomrule
  \end{tabular}
  \caption{Top 5 predicted parents by our proposed method (with PinText embeddings as input features) for a sample of test child nodes from the Pinterest knowledge graph. The true parent of each child node is in \textbf{bold}.}
  \label{tab:predictions}
\end{table*}




\subsection{Production Deployment}
\label{sec:deployment}

In three iterations prior to our proposed method being developed, knowledge graph curators at Pinterest manually expanded the Pinterest knowledge graph from $\sim$400 to $\sim$11,000 interests \citep{gonccalves2019use}. Our proposed method was subsequently deployed to help knowledge graph curators further expand the Pinterest knowledge graph from $\sim$11,000 to $\sim$28,000 interests. The decision to deploy hinged on the benchmarking results described earlier in this section.

With the same 8 knowledge graph curators used in prior knowledge graph expansion efforts, using our proposed method to provide decision support sped up the knowledge graph expansion process by $\sim$400\% on average per new node added (relative to the average time taken per node in the manual expansion immediately prior). Post expansion, a content understanding signal based on the expanded taxonomy \citep{pinterest2019pin2interest} improved the revenue of Pinterest shopping ads \citep{pinterest2022shopping} by $\sim$20\% (relative to the revenue before the expansion).

\section{Related Work}
\label{sec:related}

Our work is related to two main streams of research spanning computer science, information systems, operations, and the broader management literature, which we summarize in this section.

\textbf{Methods for knowledge graph expansion.}
Several methods have been proposed for fully-automated knowledge graph expansion in the computer science literature \citep{yu2020steam,shen2020taxoexpan,liu2021temp,wang2021enquire,zhang2021taxonomy,zeng2021enhancing,shen2022automated}. These methods essentially design increasingly performant formulations of the score function $s(u,v)$, where $u$ is the node being added to the knowledge graph and $v$ is its potential parent. Some of these methods rely on external text corpora to further improve performance. We contribute to this literature by focusing on provable human-friendliness and optimizing the performance of human-in-the-loop (or semi-automatic) knowledge graph expansion.

\textbf{Machine learning methods for human-in-the-loop decision-making.} Several recent studies have investigated how machine learning methods can be designed from the ground up to improve human decision-making. \cite{bastani2021improving} develop a reinforcement learning method to uncover interpretable ``tips'' that improve humans' sequential decision-making.  \cite{burnap2022product} propose a human-in-the-loop machine learning method to generate new product designs. \cite{ijcai2021p237} show how to improve human-algorithm collaborations with bandit feedback.
\cite{wolczynski2022learning} propose a framework to ``selectively advise'' humans, which accounts for the cost incurred by humans in ``contradicting'' predictions.
\cite{raghu2019algorithmic} show how to optimize the relative allocation of human and algorithm effort to improve the performance of their collaboration.


We contribute to this research by proposing and extensively benchmarking a human-in-the-loop machine learning method designed to improve human decision-making by limiting human effort. To the best of our knowledge, our proposed method is the first to theoretically guarantee an upper-bound on the expected human effort for both in-sample and out-of-sample data.

\textbf{The economics of human-algorithm interactions.} A stream of research analyzes the conditions under which machine learning can improve (or impair) human-in-the-loop decision-making. \cite{kleinberg2018human} show that machine learning predictions can improve judges' bail decisions when integrated into an economic framework that considers payoffs and selection biases. \cite{malik2020does} shows when and how pricing predictions based on historical human behavior can lead to increasingly inaccurate price predictions over time. \cite{balakrishnan2022} show how collaborative performance through human overrides depends on the structure of human private information and cognitive biases. \cite{grand2022best} show that machine learning methods must measure and adapt to humans' compliance with their predictions or risk poor performance. \cite{wang2022algorithmic} and \cite{mohammadi2022sell} analyze when and how algorithmic transparency and interpretability improve or worsen firm and consumer outcomes. \cite{fugener2022cognitive} study when and why humans should delegate decisions to algorithms (and vice versa).

Our work is inspired by this research but focuses on methodological contributions. Specifically, we hypothesize that our human-friendliness theoretical guarantee improves human-in-the-loop decision-making, validate our hypothesis and enhance our understanding of how it achieves this improvement with a controlled experiment, and benchmark performance on data from the field. We delegate economic analysis of the conditions required for such an improvement to future work.

\clearpage

\section{Conclusion and Limitations}
\label{sec:conclusion}

We have proposed a method for knowledge graph expansion with humans-in-the-loop. To the best of our knowledge, our proposed method is the first to \textit{provably} limit the effort required by humans to fix its predictions when incorrect. We experimentally validated that this property improves the performance of the human-algorithm collaboration by improving humans' speed and accuracy. In a case study on the Pinterest knowledge graph, we further show that our proposed method outperforms a suite of competing methods when evaluated on both accuracy and human-friendliness metrics. Our proposed method was deployed in production at Pinterest with favorable results. More generally, our work introduces an additional human-centric dimension for methodological progress --- improving humans' performance in fixing incorrect predictions.

Our work quantifies the impact of human-friendliness, but ignores the \textit{interaction} between human-friendliness and accuracy: to what extent are they substitutes, and how will this interaction impact the performance of the human-algorithm collaboration? We believe this is an important direction for future research in human-in-the-loop machine learning.

Further, we assume the availability of a candidate concepts to expand an existing knowledge graph, and do not consider the more challenging task of \textit{constructing} such a knowledge graph from scratch \citep{shen2018hiexpan,zhang2018taxogen,mao2018end}. To the best of our knowledge, knowledge graph construction with humans-in-the-loop remains an open problem, which we believe is an important direction for future research on knowledge graphs.

Theoretically, our work provides an \textit{average}-case guarantee for the human effort required by our proposed method, assuming independent and identically distributed training data. We believe that extending our theory to provide \textit{worst}-case guarantees (by upper-bounding the \textit{maximum} human effort) and relaxing our assumption of independent and identically distributed training data are good candidates for future theoretical research with a potentially high practical impact.

Finally, we have not considered several other important dimensions of human-centric machine learning: incorporating human judgement and feedback \citep{ibrahim2021eliciting,ijcai2021p237}, quantifying and depicting algorithmic uncertainty \citep{mcgrath2020does}, exhibiting fairness \citep{fu2020artificial,de2022algorithmic}, and being interpretable and transparent \citep{smith2020no}. We believe incorporating each of these dimensions is worthy of future research.

\bibliographystyle{emaad}
\bibliography{scibib}

\clearpage

\section*{Appendix A: Statistical Significance with Decision Support Correctness}

\begin{table*}[t]
  \centering
  \begin{tabular}{l@{\hspace{2.5em}}c@{\hspace{1.5em}}c@{\hspace{1.5em}}c@{\hspace{1.5em}}c}
  \toprule
  & \multicolumn{4}{c}{Dependent Variable}\\
  \cmidrule(lr){2-5}
  &  & Time per & Decision  &  \\
  & Total Score & Prompt & Accuracy (\%) & Compliance (\%) \\
  & (1) & (2) & (3) & (4)\\
  \midrule
  $\mathbb{I}[\textrm{Not-Human-Friendly}]$              & $-13.09^{***}$ & $8.39^{***}$  & $-22.09^{***}$ & $-3.22^{*}$  \\
                                                         & $(1.62)$       & $(1.50)$      & $(2.83)$      & $(1.25)$     \\[0.5em]
  $\mathbb{I}[\textrm{Decison Support Correct}]$         & $8.38^{***}$   & $-1.43^{*}$   & $7.11^{***}$  & $74.77^{***}$ \\
                                                         & $(1.19)$       & $(0.58)$      & $(1.91)$      & $(1.82)$     \\[0.5em]
  $\mathbb{I}[\textrm{Not-Human-Friendly}] \times$       & $7.56^{***}$   & $-7.76^{***}$ & $21.33^{***}$  & $2.45$       \\
  \quad$\mathbb{I}[\textrm{Decison Support Correct}]$    & $(1.82)$       & $(0.83)$      & $(2.75)$      & $(2.35)$     \\[0.5em]
  Intercept                                              & $14.66^{***}$  & $16.52^{***}$ & $75.23^{***}$  & $7.57^{***}$ \\
                                                         & $(1.30)$       & $(0.95)$      & $(1.89)$      & $(1.02)$     \\
  \midrule
  $R^2$                                                  & $0.44$         & $0.23$        & $0.48$        & $0.95$       \\
  Adjusted $R^2$                                         & $0.43$         & $0.22$        & $0.47$        & $0.95$       \\
  $N$                                                    & $208$          & $208$         & $208$         & $208$        \\
  \bottomrule
  \end{tabular}
  \\[0.5em]
  \begin{small}
    Note: Robust standard errors clustered by subject are in parentheses. $^{***}p<0.001$;$^{**}p<0.01$;$^{*}p<0.05$
  \end{small}
  \caption{Estimates from linear regressions of various dependent variables (conditional on decision support correctness) on the treatment condition indicator interacted with the decision support correctness indicator.}
  \label{tab:expreg}
\end{table*}

In Section \ref{sec:results}, we quantified differences in the total score, average decision time per prompt, decision accuracy, and compliance with the decision support for subjects in the Human-Friendly (HF) and Not-Human-Friendly (NHF) treatment conditions after conditioning on the decision support correctness. In this section, we assess the statistical significance of these differences.

Specifically, for each of the 4 aforementioned dependent variables, we construct 2 separate versions of the dependent variable for each subject: one conditional on prompts where the decision support is correct, and the other conditional on prompts where the decision support is incorrect. For example, we calculate 2 total scores for each subject: one that only includes those prompts where the decision support is correct, and the other that only includes those prompts where the decision support is incorrect.

We then regress these dependent variables on (i) an indicator for the Not-Human-Friendly (NHF) treatment condition, (ii) a binary indicator that equals 1 when the dependent variable is conditional on the decision support being correct, and equals 0 when the dependent variable is conditional on the decision support being incorrect, and (iii) the interaction of the two indicators described in (i) and (ii).
The results from each of these regressions are in Table \ref{tab:expreg}. In Table \ref{tab:expAMEs}, we also report the average marginal effects of $\mathbb{I}[\textrm{Not-Human-Friendly}]$ corresponding to each of the regressions in Table \ref{tab:expreg}. Robust standard errors clustered by subject are in parentheses.

\begin{table*}
  \centering
  \begin{tabular}{l@{\hspace{2.5em}}c@{\hspace{1.5em}}c@{\hspace{1.5em}}c@{\hspace{1.5em}}c}
  \toprule
  & \multicolumn{4}{c}{Dependent Variable}\\
  \cmidrule(lr){2-5}
  &  & Time per & Decision  &  \\
  & Total Score & Prompt & Accuracy (\%) & Compliance (\%) \\
  & (1) & (2) & (3) & (4)\\
  \midrule
  $\mathbb{I}[\textrm{Decison Support Incorrect}]$ & $-13.09^{***}$ & $8.39^{***}$           & $-22.1\%^{***}$ & $-3.2\%^{**}$    \\
  & (1.62)         & (1.49)                 & (2.8\%)         & (1.2\%)          \\[0.5em]
  $\mathbb{I}[\textrm{Decison Support Correct}]$ & $-5.52 ^{**}$  & 0.63                   & -0.8\%          & -0.8\%           \\
  & (1.99)         & (1.42)                 & (2.0\%)         & (2.1\%)          \\
  \bottomrule
  \end{tabular}
  \\[0.5em]
  \begin{small}
    Note: Robust standard errors clustered by subject are in parentheses. $^{***}p<0.001$;$^{**}p<0.01$;$^{*}p<0.05$
  \end{small}
  \caption{Estimates of the average marginal effect of $\mathbb{I}[\textrm{Not-Human-Friendly}]$ for the regressions in Table \ref{tab:expreg}.}
  \label{tab:expAMEs}
\end{table*}

\section*{Appendix B: Implementation Details}

All our benchmarking was performed on a cluster of 4 machines, each with two 24-core 3GHz Intel Xeon Gold 6248R CPUs, 512GB of RAM, and an Nvidia Tesla T4 GPU with 16GB of VRAM.

We adopted a few training tricks from word2vec \citep{mikolov2013distributed}. We normalized all feature vectors to have unit $L_2$ norm. For our proposed method, we created separate copies of the feature vector of each node to use when the node appears as a child and as a parent. We only freeze the child copy during training. All model parameters were initialized randomly.

We tuned the hyperparameters of all benchmarked methods via grid search on the validation dataset. Batch sizes in each grid were selected based on the available VRAM (16GB). We trained each benchmarked method until the validation MRR showed no increase for 10 training epochs (or iterations in the case of the fine-tuned transformers). In Table \ref{tab:hyperparameters}, we report the hyperparameter grids and the selected hyperparameter values for our proposed method and all baselines.

We use the model stored at the epoch with the highest validation MRR for the final evaluation on the test dataset (we found that using the lowest validation loss instead lead to poorer performance).

\begin{table*}[t]
  \centering
  \begin{tabular}{l@{\hspace{1.5em}}l@{\hspace{1.5em}}c}
    \toprule
    \textbf{Method} & \textbf{Hyperparameter Grid} & \textbf{Value}\\
    \midrule
    Feedforward Neural Network & Batch size $\in \{2^{10}, 2^{11}, 2^{12}, 2^{13}\}$ & $2^{12}$ \\
    & Learning rate $\in \{10^{-4}, 10^{-3}, 10^{-2}, 10^{-1}, 10^0\}$ & $10^{-4}$ \\
    & Weight decay $\in \{10^{-4}, 10^{-3}, 10^{-2}, 10^{-1}, 10^0\}$ & $10^{0}$ \\
    & $f(\cdot)$ hidden layer size $\in \{100, 150, 200, 250, 300\}$ & 150 \\
    & $f(\cdot)$ number of hidden layers $\in \{1, 2, 3, 4\}$ & 2 \\
    & Activation $\in$ \{ReLU, TanH, Sigmoid\} & ReLU\\
    & Optimization method $\in$ \{Adam, AdamW, SGD\} & Adam\\
    \midrule
    LambdaMART & Number of estimators $\in \{10^1, 10^2, 10^3\}$ & $10^3$\\
    & Number of leaves $\in \{2^5-1, 2^8-1, 2^{11} - 1\}$ & $2^8 -1$ \\
    & Learning rate $\in \{10^{-4}, 10^{-3}, 10^{-2}, 10^{-1}, 10^0\}$ & $10^{-1}$ \\
    & Regularization $\in \{10^{-4}, 10^{-3}, 10^{-2}, 10^{-1}, 10^0\}$ & $10^{-4}$ \\
    \midrule
    MiniLM-L12-H384-uncased & Learning rate $\in \{10^{-7}, 10^{-5}, 10^{-3}, 10^{-1}\}$ & $10^{-5}$ \\
    & Weight decay $\in \{10^{-7}, 10^{-5}, 10^{-3}, 10^{-1}\}$ & $10^{-7}$\\
    \midrule
    all-miniLM-L6-v2 & Learning rate $\in \{10^{-7}, 10^{-5}, 10^{-3}, 10^{-1}\}$ & $10^{-5}$ \\
    & Weight decay $\in \{10^{-7}, 10^{-5}, 10^{-3}, 10^{-1}\}$ & $10^{-7}$\\
    \midrule
    msmarco-miniLM-L6-v2 & Learning rate $\in \{10^{-7}, 10^{-5}, 10^{-3}, 10^{-1}\}$ & $10^{-5}$ \\
    & Weight decay $\in \{10^{-7}, 10^{-5}, 10^{-3}, 10^{-1}\}$ & $10^{-7}$\\
    \midrule
    Triplet Matching Network & $k \in \{5, 10, 15, 20\}$ & $10$ \\
    & Batch size $\in \{2^{6}, 2^{7}, 2^{8}, 2^{9}\}$ & $2^{8}$ \\
    & Learning rate $\in \{10^{-4}, 10^{-3}, 10^{-2}, 10^{-1}, 10^0\}$ & $10^{-2}$ \\
    & Weight decay $\in \{10^{-4}, 10^{-3}, 10^{-2}, 10^{-1}, 10^0\}$ & $10^{-4}$ \\
    & Optimization method $\in$ \{Adam, AdamW, SGD\} & Adam\\
    \midrule
    Proposed Method & $k \in \{2^4, 2^5, 2^6, 2^7\}$ & $2^7$ \\
                    & Batch size $\in \{2^{10}, 2^{11}, 2^{12}, 2^{13}\}$ & $2^{13}$ \\
                    & Learning rate $\in \{10^{-4}, 10^{-3}, 10^{-2}, 10^{-1}, 10^0\}$ & $10^{-3}$ \\
                    & Weight decay $\in \{10^{-4}, 10^{-3}, 10^{-2}, 10^{-1}, 10^0\}$ & $10^{0}$ \\
                    & $f(\cdot)$ hidden layer size $\in \{100, 150, 200, 250, 300\}$ & 150 \\
                    & $f(\cdot)$ number of hidden layers $\in \{1, 2, 3, 4\}$ & 2 \\
                    & Optimization method $\in$ \{Adam, AdamW, SGD\} & AdamW\\
    \bottomrule
  \end{tabular}
  \caption{Hyperparameter grids and selected hyperparameter values for all benchmarked methods.}
  \label{tab:hyperparameters}
\end{table*}

%

\end{document}